\documentclass[10pt,twocolumn,letterpaper]{article}

\usepackage{cvpr}              % To produce the CAMERA-READY version
\definecolor{cvprblue}{rgb}{0.21,0.49,0.74}
\usepackage[pagebackref,breaklinks,colorlinks,allcolors=cvprblue]{hyperref}

\usepackage{url}
\usepackage{graphicx} % Optional: improves spacing with \textwidth
\usepackage{booktabs}       % professional-quality tables
\usepackage{pifont}
\usepackage{multirow}
\usepackage{wrapfig}

\def\paperID{*****} % *** Enter the Paper ID here
\def\confName{CVPR}
\def\confYear{2026}

\title{Shoe Style-Invariant and Ground-Aware \\ Learning for Dense Foot Contact Estimation}

\author{
  Daniel Sungho Jung$^{1}$ \hskip1.6em Kyoung Mu Lee$^{1,2}$ \\
   $^{1}$IPAI, $^{2}$Dept. of ECE \& ASRI, Seoul National University, Korea  \\ 
   {\tt\small \{dqj5182, kyoungmu\}@snu.ac.kr} 
}

\begin{document}
\maketitle

\begin{abstract}
Foot contact plays a critical role in human interaction with the world, and thus exploring foot contact can advance our understanding of human movement and physical interaction. 
Despite its importance, existing methods often approximate foot contact using a zero-velocity constraint and focus on joint-level contact, failing to capture the detailed interaction between the foot and the world. 
Dense estimation of foot contact is crucial for accurately modeling this interaction, yet predicting dense foot contact from a single RGB image remains largely underexplored.
There are two main challenges for learning dense foot contact estimation.
First, shoes exhibit highly diverse appearances, making it difficult for models to generalize across different styles.
Second, ground often has a monotonous appearance, making it difficult to extract informative features.
To tackle these issues, we present a FEet COntact estimation (FECO) framework that learns dense foot contact with shoe style-invariant and ground-aware learning. 
To overcome the challenge of shoe appearance diversity, our approach incorporates shoe style adversarial training that enforces shoe style-invariant features for contact estimation. 
To effectively utilize ground information, we introduce a ground feature extractor that captures ground properties based on spatial context. 
As a result, our proposed method achieves robust foot contact estimation regardless of shoe appearance and effectively leverages ground information. 
The codes are available at \url{https://github.com/dqj5182/FECO_RELEASE}.
\end{abstract}
\section{Introduction}

Human movement and balance fundamentally depend on how the feet interact with the surrounding environment. 
Every step, shift in posture, or change in direction involves complex physical exchanges between the foot and the ground or an object. 
Understanding these interactions is essential for interpreting human motion, maintaining stability, and modeling realistic physical behavior. 
By accurately capturing where the foot establishes contact with the environment, systems can reason about human interaction dynamics from visual input, enabling more realistic and physically consistent human behavior perception.

Despite its importance, existing works predominantly simplify foot contact into joint-level contact~\cite{zou2020reducing, rempe2021humor, mourot2022underpressure, yi2022physical, zhuo2023towards, shin2024wham} and often rely on geometric heuristics such as zero-velocity constraints~\cite{zou2020reducing, shin2024wham}. 
These formulations fail to capture the dense and spatially distributed foot contact, where the contact region is distributed across multiple fine-grained regions of the foot. 
Although several dense body contact estimation methods~\cite{hassan2021populating, huang2022capturing, tripathi2023deco, nam2024joint} have been proposed, their foot contact predictions remain inaccurate.
Recently, a study on hand contact~\cite{jung2025learning} showed that a dedicated dense contact estimation model for a specific body part can outperform general dense body contact models.
Motivated by this, we aim to develop a dedicated dense foot contact estimation model trained on a large-scale dataset to advance the understanding of human movement and interaction.

Estimating dense foot contact is particularly challenging due to two crucial factors: appearance diversity and ground ambiguity.
In the real world, feet are typically covered with shoes, which exhibit immense variation in color, texture, material, and style.
Such diversity introduces spurious visual factors that correlate with contact patterns in the training data but are unrelated to true physical interaction, often causing models to rely on misleading appearance cues rather than geometric reasoning.
For instance, in a training dataset, individuals wearing sneakers may frequently perform actions such as abrupt walking or skateboarding, leading the model to associate specific shoe styles with certain contact patterns.
This bias can cause the model to overfit to appearance-dependent correlations and perform poorly when encountering unseen foot actions for the shoe type.
At the same time, ground surfaces such as carpet, asphalt, or polished floors often have weak or repetitive textures, providing limited visual cues for inferring contact regions.
Yet, understanding the ground is critical for foot–ground contact estimation, since contact inherently occurs along the direction parallel to the surface.
Without explicit reasoning about ground geometry, models fail to identify the support level and consequently produce inaccurate contact predictions.
These difficulties are further compounded by occlusion, viewpoint changes, and illumination variation, all of which highlight the need for representations that capture geometric and physical context rather than superficial ground appearance.
Therefore, building a robust dense foot contact estimation model requires both strong shoe style invariance and a grounded understanding of the physical surface.

To address these issues, we introduce FECO, a framework for dense foot contact estimation that learns shoe style-invariant and ground-aware representations from a single image. FECO begins with a low-level style randomization stage using progressive random convolutions~\cite{choi2023progressive} to eliminate reliance on local and low-level texture statistics. 
We then perform shoe style–content randomization using the external shoe image dataset UT Zappos50K~\cite{yu2014fine}.
In this process, adversarial adapters in the shoe content randomization branch enforce invariance to shoe style, while the shoe style randomization branch perturbs shoe appearances during training to expose the model to diverse visual styles.
This dual randomization strategy effectively disentangles style from content, enabling the model to focus on structural cues relevant to dense foot contact and improving robustness to unseen shoe types and materials.
Complementing style-invariant learning, FECO explicitly models the ground through pixel height maps~\cite{sheng2022controllable} and ground normals. 
These capture geometric and spatial cues that correlate with ground, even for visually ambiguous surfaces.

As a result, FECO achieves state-of-the-art performance across diverse shoe and ground appearance with its dedicated shoe style-invariant and ground-aware learning for dense foot contact estimation.

Our key contributions are as follows:
\begin{itemize}
\item We introduce FECO, a novel framework for dense foot contact estimation from a single image that explicitly targets shoe style diversity issue and foot-ground reasoning issue and allows robust contact prediction.
\item To mitigate shoe style diversity issue, we propose shoe style–content randomization, which enforces style-invariant yet content-preserving features using an external shoe dataset for robustness towards shoe style diversity.
\item To reason foot-ground contact, we present ground-aware learning that encodes geometric properties of the ground with supervision from pixel height maps and ground normals, enhancing foot-ground contact reasoning.
\item In the end, FECO, the first method for a dedicated dense foot contact estimation demonstrates strong performance across diverse benchmarks.
\end{itemize}

\begin{table}[t]
\centering
\caption{\textbf{Dataset configuration.} We leverage 10 datasets with diverse foot contact.}
\vspace{-0.5em}
\scalebox{0.83}{\begin{tabular}{lccc} \toprule
Dataset & Interaction & Label & \# of images \\
\midrule
PROX~\cite{hassan2019resolving} & Scene & 3D Mesh & 0.2K \\
BEHAVE~\cite{bhatnagar2022behave} & Ground + Object & 3D Mesh & 45K \\
InterCap~\cite{huang2022intercap} & Ground + Object & 3D Mesh & 61K \\
EgoBody~\cite{zhang2022egobody} & Scene & 3D Mesh & 220K \\
RICH~\cite{huang2022capturing} & Scene & 3D Mesh & 540K \\
MOYO~\cite{tripathi20233d} & Ground & Pressure Mat & 560K \\
Hi4D~\cite{yin2023hi4d} & Ground + Body & 3D Mesh & 11K \\
MMVP~\cite{zhang2024mmvp} & Ground & Insole & 44K \\
MotionPRO~\cite{ren2025motionpro} & Ground & Pressure Mat & 12.4M \\
\midrule
COFE dataset & Scene & 2D Keypoint & 31K \\

\bottomrule
\end{tabular}}
\label{tab:dataset}
\end{table}
\section{Related works}

\paragraph{Foot contact estimation.}
Foot contact estimation has been studied through two main approaches, joint-level foot contact estimation and dense foot contact estimation as part of dense human body contact estimation. 
In the joint-level setting, Footskate Reducer~\cite{zou2020reducing} formulated foot contact using a zero-velocity constraint, where a foot joint is considered to be in contact if its 2D displacement between consecutive frames is below a threshold.
HuMoR~\cite{rempe2021humor} jointly trained human-ground contact and 3D human motion with ground-truth joint-level contact extracted with fitted ground and 3D human joints.
UnderPressure~\cite{mourot2022underpressure} presented a framework that predicts vertical ground reaction forces from sequences of 3D joints to extract foot joint contact.
PIP~\cite{yi2022physical} estimated and utilized foot-ground contact as intermediate prediction for requiring ground reaction force.
Foot Stabilization~\cite{zhuo2023towards} introduced pseudo-GT annotator for foot-ground contacts by thresholding the distance between SMPL meshes and ground plane.
WHAM~\cite{shin2024wham} utilized the zero-velocity constraint to extract foot contact, which serves as main cue for refining 3D global trajectory.
In the dense contact setting, POSA~\cite{hassan2021populating} learned dense contact by conditioning 3D vertex position on conditional variational autoencoder (cVAE)~\cite{sohn2015learning}.
BSTRO~\cite{huang2022capturing} directly estimated dense contact from visual input through Transformer-based architecture~\cite{devlin2019bert}.
DECO~\cite{tripathi2023deco} introduced novel annotation technique for labeling contact on in-the-wild images.

\paragraph{Reducing style bias for generalization.}
Style is arguably one of the core representations that expresses an image.
In early works, image style has been widely studied in a task of image style transfer~\cite{gatys2016image} that separates and recombines low-level image style and content to new images using Gram matrix~\cite{gatys2015texture}.
Motivated by the literature, StyleGAN~\cite{karras2019style} explored the GAN-based architecture that learns to effectively separate, control, and interpolate on high-level style attributes when generating an image.
Moving on to the discriminative tasks, there have been extensive studies on how to effectively handle image style to improve performance.
BIN~\cite{nam2018batch} leveraged Batch Normalization~\cite{ioffe2015batch} and Instance Normalization~\cite{ulyanov2016instance} with gating mechanism to learn which style to use across different task selectively from individual feature maps.
SagNets~\cite{nam2021reducing} built two separate networks of content-biased network and style-biased network to learn content while adversarially learn to reduce dependency on low-level style with style and content randomization.
RandConv~\cite{xurobust} proposed to construct infinite number of new domains by applying random convolution filter to image before feeding to image encoder, which allows training on random local texture. 
ABA~\cite{cheng2023adversarial} trained Bayesian Neural Network to augment image that serves as the most effective adversarial example for classifier to boost domain generalization.
StyDeSty~\cite{liu2024stydesty} introduced stylization and destylization module that builds domain-invariant representation to guarantee the alignment between augmented domain and target domain.
To use diverse high-level and low-level shoe styles from UT Zappos50K~\cite{yu2014fine} dataset, we employ a variant of SagNets for our shoe style-invariant learning.

\paragraph{Ground representation.}
Understanding ground information has been studied in the field of autonomous driving~\cite{zhou2021monocular, qin2022monoground, moon2024ground, cecille2024groco} and depth estimation~\cite{yang2023gedepth, moon2024ground, cecille2024groco} to use ground as reference geometry for overall prediction.
Mono3D~\cite{chen2016monocular} set a heuristic prior that objects are always on the ground and place proposals for 3D bounding box on the ground.
% MonoEF~\cite{zhou2021monocular}
Ground-Aware~\cite{liu2021ground} proposed a ground-aware convolution module that utilizes each pixel's prior depth value and features from pixels below the target pixel to guide ground-based reasoning.
MonoGround~\cite{qin2022monoground} assumed that the bottom of object's 3D bounding box as ground plane and leveraged the information to extract projected dense ground depth.
% GEDepth~\cite{yang2023gedepth}
FGTO~\cite{moon2024ground} also made assumption that dynamic objects tend to be in contact with the ground, and proposed GDS loss that aligns depth of the dynamic objects and ground in self-supervised manner.
GroCo~\cite{cecille2024groco} proposed to predict accurate ground attention map that is utilized to guide refinement of predicted depth map.
However, most of the works relied on pre-determined camera extrinsic from the vehicle to extract ground information.
In the literature of shadow generation, a novel representation called Pixel Height~\cite{sheng2022controllable} was introduced to model correlation between object and ground.
The Pixel Height representation was also proved effective on light effect generation by PixHt-Lab~\cite{sheng2023pixht} where the paper utilized pixel height map for object and ground to reconstruct 3D model.
Recently, ORG~\cite{man2025floating} extended the use of Pixel Height to encode object-ground relationship and make the reconstructed 3D geometry to be aligned with ground.
Our FECO also extends this literature of Pixel Height but on foot contact estimation task for learning foot-ground interaction in fine-grained pixel-level representation.
\section{Method: FECO}
\begin{figure*}[t]
\begin{center}
\includegraphics[width=1.0\linewidth]{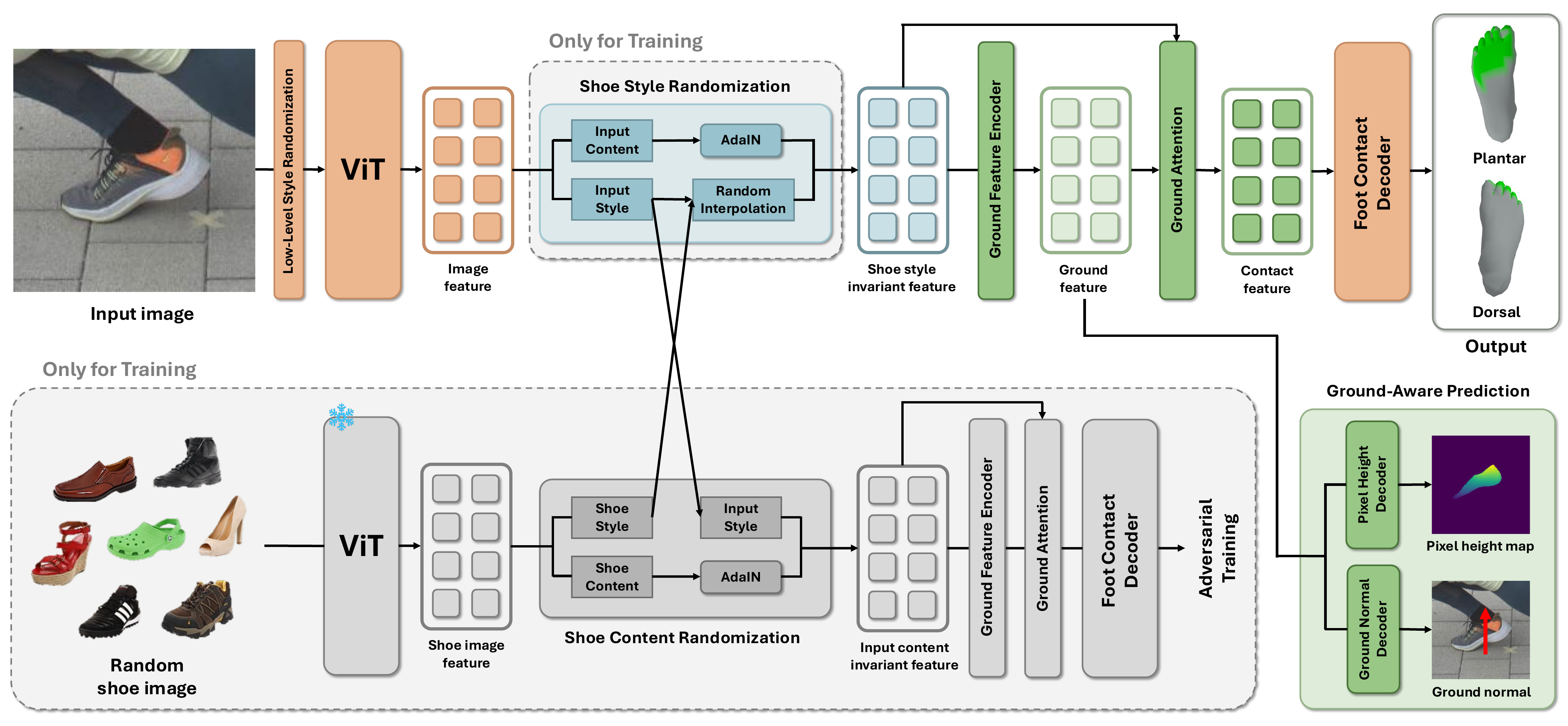}
\end{center}
\vspace{-4mm}
\caption{
\textbf{Overall pipeline of FECO.} Our method first applies low-level style randomization on input image and encodes it into image feature using a ViT backbone.
From image feature, shoe style and shoe content randomization are performed with random shoe images from the UT Zappos50K~\cite{yu2014fine} dataset to produce a shoe style-invariant feature.
This feature is then processed by a ground feature encoder to extract ground feature, which is used to predict pixel height map and ground normal.
Finally, the ground feature and shoe style-invariant feature are fused to form a contact feature, which is decoded to produce the final foot contact prediction.}
\label{fig:pipeline}
\vspace{-0.3cm}
\end{figure*}

Figure~\ref{fig:pipeline} shows the overall pipeline of FECO, which consists of low-level style randomization, shoe style–content randomization, ground feature learning, spatial attention, and foot contact decoder. 
The network is trained with three synchronized inputs per sample, namely one clean image and two low-level style randomized images, and we train all modules in an end-to-end manner.

\subsection{Low-level style randomization}
Given an RGB image~$\mathbf{I} \in \mathbb{R}^{3 \times H \times W}$, where $H$ and $W$ represent the spatial resolution, we first construct three separate images that consists of one clean image~$\mathbf{I}_{\mathrm{lr}}^{(0)}$ and two augmented images~$\mathbf{I}_{\mathrm{lr}}^{(1)}$, $\mathbf{I}_{\mathrm{lr}}^{(2)}$ produced by Pro-RandConv~\cite{choi2023progressive}.
Following Pro-RandConv, we randomly sample the convolution weights $w$, the deformable convolution offset $\Delta p$, and the affine parameters $\gamma$ and $\beta$ independently for each of the two augmentation for producing images~$\mathbf{I}_{\mathrm{lr}}^{(1)}$, $\mathbf{I}_{\mathrm{lr}}^{(2)}$. 
Then, using the sampled parameters, we apply deformable convolution followed by per-sample, per-channel standardization with Instance Normalization~\cite{ulyanov2016instance}.
Afterward, an affine transformation with parameters $\gamma$ and $\beta$ is applied, followed by a hyperbolic tangent, which ultimately produces the low-level style randomized images $\mathbf{I}_{\mathrm{lr}}^{(k)}$ for $k \in \{1,2\}$. 
As a result, we obtain one clean image and two low-level style randomized images, which we denote collectively as $\mathbf{I}_{\mathrm{lr}}^{(k)}$ for $k \in \{0,1,2\}$. 
During training, all three images are processed with identical operations, while at inference only the clean image is considered.
For clarity in the following sections, we simply write $\mathbf{I}_{\mathrm{lr}}$ to represent the images.

\subsection{Shoe style-content randomization}
\label{sec:shoe_style_contact_rand}
The shoe style-content randomization is designed to make FECO robust on random shoe style and make it focus more on content of the input image.
We split into two pipeline to conduct shoe style randomization and shoe content randomization that each randomize shoe style during the training of dense foot contact estimation, and conduct adversarial training on shoe content-randomized feature.
Before conducting shoe style-content randomization, we first resize the input image~$\mathbf{I}_{\mathrm{lr}}$ with bilinear interpolation and extract an image feature $\mathbf{F} \in \mathbb{R}^{c \times h \times w}$ with a Vision Transformer~(ViT)~\cite{dosovitskiy2020image} from resized image~$\mathbf{I}_{\mathrm{lr}} \in \mathbb{R}^{3 \times 224 \times 224}$. 
A foot segmentation mask $\mathbf{M}_{\mathrm{f}} \in \mathbb{R}^{h \times w}$ is predicted from image feature~$\mathbf{F}$ using a DPT decoder~\cite{ranftl2021vision}.
The foot mask~$\mathbf{M}_{\mathrm{f}}$ is used for foot localized operations.
A random shoe image $\mathbf{I}_{\mathrm{s}}$ from a dedicated shoe dataset~\cite{yu2014fine} is also encoded with a frozen ViT backbone to obtain a shoe feature $\mathbf{F}_{\mathrm{s}} \in \mathbb{R}^{c \times h \times w}$. 
Unlike SagNets~\cite{nam2021reducing}, which select a style source from other samples within mini-batch, our approach uses an external shoe dataset to ensure diverse and independent style sources.

\noindent\textbf{Shoe content randomization.}
We aim to construct a shoe style biased feature that preserves the content of the shoe feature~$\mathbf{F}_{\mathrm{s}}$ while adopting the style of the input feature~$\mathbf{F}$. 
To stabilize adversarial training, we introduce two adapters $\mathbf{A}_{\mathrm{prev}}$ and $\mathbf{A}_{\mathrm{after}}$, each implemented as a $3 \times 3$ convolution with zero-initialized weights and scaled by a learnable parameter $\gamma$ initialized to $0.02$. 
In residual form, an adapter operates as
\begin{equation}
\mathbf{A}(\mathbf{X}) = \mathbf{X} + \gamma \, \phi(\mathbf{X}),
\label{eq:adapter_form}
\end{equation}
where $\phi$ denotes the convolutional mapping. 
We apply $\mathbf{A}_{\mathrm{prev}}$ to both input and shoe features, yielding $\mathbf{F}' = \mathbf{A}_{\mathrm{prev}}(\mathbf{F})$ and $\mathbf{F}'_{\mathrm{s}} = \mathbf{A}_{\mathrm{prev}}(\mathbf{F}_{\mathrm{s}})$. 
The channel-wise statistics are then computed as
\begin{equation}
\begin{aligned}
\mu' &= \mu(\mathbf{F}'), \quad \sigma' = \sigma(\mathbf{F}'), \\
\mu'_{\mathrm{s}} &= \mu(\mathbf{F}'_{\mathrm{s}}), \quad \sigma'_{\mathrm{s}} = \sigma(\mathbf{F}'_{\mathrm{s}}).
\end{aligned}
\label{eq:content_stats}
\end{equation}
Using $(\mu', \sigma')$ as the target style, we normalize the shoe feature with its own statistics $(\mu'_{\mathrm{s}}, \sigma'_{\mathrm{s}})$ and re-scale it as
\begin{equation}
\mathbf{F}_{\mathrm{ici}} = \mathbf{A}_{\mathrm{after}} \!\left( \sigma' \cdot \frac{\mathbf{F}'_{\mathrm{s}} - \mu'_{\mathrm{s}}}{\sigma'_{\mathrm{s}}} + \mu' \right),
\label{eq:final_ici}
\end{equation}
where $\mathbf{F}_{\mathrm{ici}}$ is input content invariant feature.
The resulting feature~$\mathbf{F}_{\mathrm{ici}}$ preserves the content of the external shoe feature while expressing it in the style of the input. 
This content-randomized branch is explicitly used for adversarial training, ensuring that the final predictor does not overfit to style cues from the input image.

\noindent\textbf{Shoe style randomization.}
In parallel, we construct a complementary shoe style-invariant representation. 
Using two additional adapters for alignment, we apply $\mathbf{A}_{\mathrm{prev}}$ to both input and shoe features, yielding $\mathbf{F}' = \mathbf{A}_{\mathrm{prev}}(\mathbf{F})$ and $\mathbf{F}'_{\mathrm{s}} = \mathbf{A}_{\mathrm{prev}}(\mathbf{F}_{\mathrm{s}})$. 
A random interpolation weight $\alpha$ is drawn from a uniform distribution, and the interpolated statistics are defined as
\begin{equation}
\begin{aligned}
\hat{\mu} &= \alpha \cdot \mu(\mathbf{F}') + (1-\alpha) \cdot \mu(\mathbf{F}'_{\mathrm{s}}), \\
\hat{\sigma} &= \alpha \cdot \sigma(\mathbf{F}') + (1-\alpha) \cdot \sigma(\mathbf{F}'_{\mathrm{s}}).
\end{aligned}
\label{eq:interp_style}
\end{equation}
AdaIN~\cite{huang2017arbitrary} is then applied to $\mathbf{F}'$ with these interpolated statistics, and $\mathbf{A}_{\mathrm{after}}$ refines the results, producing the shoe style-invariant representation
\begin{equation}
\mathbf{F}_{\mathrm{ssi}} = \mathbf{A}_{\mathrm{after}} \!\left( \hat{\sigma} \cdot \frac{\mathbf{F}' - \mu(\mathbf{F}')}{\sigma(\mathbf{F}')} + \hat{\mu} \right),
\label{eq:final_ssi}
\end{equation}
where $\mathbf{F}_{\mathrm{ssi}}$ is shoe style-invariant feature.
In the end, this branch incorporates style from external shoe image features to accomplish shoe style randomization on input image feature during training.

\subsection{Ground feature learning}
To enhance the model’s understanding of the ground, the most frequent contact surface for the foot, we introduce ground feature learning with two supervisory signals: pixel height maps~$\mathbf{PH}$ and ground normals~$\mathbf{N}_{\mathrm{g}}$. 
This learning is applied in parallel to both input content invariant features~$\mathbf{F}_{\mathrm{ici}}$ and shoe style-invariant features~$\mathbf{F}_{\mathrm{ssi}}$.

\noindent\textbf{Ground feature encoding.}
A ground feature encoder~$\mathrm{Enc}_{\mathrm{g}}$ processes each randomized feature independently, producing multi-level representations
\begin{equation}
\begin{aligned}
\{\mathbf{F}_{\mathrm{g},l}^{\mathrm{ici}}\}_{l=1}^4 &= \mathrm{Enc}_{\mathrm{g}}(\mathbf{F}_{\mathrm{ici}}), \\
\{\mathbf{F}_{\mathrm{g},l}^{\mathrm{ssi}}\}_{l=1}^4 &= \mathrm{Enc}_{\mathrm{g}}(\mathbf{F}_{\mathrm{ssi}}).
\end{aligned}
\label{eq:ground_enc}
\end{equation}
Each encoder level $l$ consists of a $3 \times 3$ convolution with ReLU activation, followed by a $1 \times 1$ convolution with another ReLU. 
The multi-level features are designed to capture fine-grained information for pixel height map decoding.
For ground normal decoding, we only utilize the final-level feature due to its global nature.

\noindent\textbf{Pixel height decoding.}
From the multi-level ground features, a DPT decoder~\cite{ranftl2021vision} predicts dense pixel height maps,
\begin{equation}
\begin{aligned}
\mathbf{PH}^{\mathrm{ici}} &= \mathrm{Dec}_{\mathrm{ph}}(\{\mathbf{F}_{\mathrm{g},l}^{\mathrm{ici}}\}_{l=1}^4), \\
\mathbf{PH}^{\mathrm{ssi}} &= \mathrm{Dec}_{\mathrm{ph}}(\{\mathbf{F}_{\mathrm{g},l}^{\mathrm{ssi}}\}_{l=1}^4).
\end{aligned}
\label{eq:ph_dec}
\end{equation}
The decoder head outputs a single channel per spatial location, and predictions are upsampled to the input resolution.
Motivated by depth scaling factor from Depth Anything~\cite{yang2024depth}, we scale the predicted height maps by the maximum side length $s = \max(H, W)$ of the input image so that the outputs are expressed in pixel units.

\noindent\textbf{Ground normal decoding.}
The ground normal in camera coordinates is predicted from the final-level ground features~($\mathbf{F}_{\mathrm{g},4}^{\mathrm{ici}}$ or $\mathbf{F}_{\mathrm{g},4}^{\mathrm{ssi}}$). 
To prevent shortcut learning from the orientation of the foot, we suppress the foot region using the foot segmentation mask $\mathbf{M}_{\mathrm{f}}$, yielding
\begin{equation}
\begin{aligned}
\mathbf{F}_{\mathrm{g}}^{\mathrm{m,ici}} &= \mathbf{F}_{\mathrm{g},4}^{\mathrm{ici}} \odot (1-\mathbf{M}_{\mathrm{f}}), \\
\mathbf{F}_{\mathrm{g}}^{\mathrm{m,ssi}} &= \mathbf{F}_{\mathrm{g},4}^{\mathrm{ssi}} \odot (1-\mathbf{M}_{\mathrm{f}}).
\end{aligned}
\label{eq:masked_ground_features}
\end{equation}
Our ground normal decoder~$\mathrm{Dec}_{\mathrm{gn}}$ applies global average pooling, two fully connected layers with a hidden dimension of 128, a $\tanh$ activation, and $\ell_2$ normalization. 
Applying it to each branch gives
\begin{equation}
\begin{aligned}
\mathbf{N}_{\mathrm{g}}^{\mathrm{ici}} &= \mathrm{Dec}_{\mathrm{gn}}(\mathbf{F}_{\mathrm{g}}^{\mathrm{m,ici}}), \\
\mathbf{N}_{\mathrm{g}}^{\mathrm{ssi}} &= \mathrm{Dec}_{\mathrm{gn}}(\mathbf{F}_{\mathrm{g}}^{\mathrm{m,ssi}}).
\end{aligned}
\label{eq:ground_normals}
\end{equation}
which are unit-length ground normal vectors.

\subsection{Spatial attention}
We fuse the randomized features~($\mathbf{F}_{\mathrm{ici}}$ and $\mathbf{F}_{\mathrm{ssi}}$) with their corresponding ground features~($\mathbf{F}_{\mathrm{g}}^{\mathrm{ici}}$ and $\mathbf{F}_{\mathrm{g}}^{\mathrm{ssi}}$) using a spatial attention module. 
For each randomization branch, the randomized and ground features are concatenated along the channel dimension and passed through a $3 \times 3$ convolution to reduce the channel dimension to 256, followed by ReLU activation and a dropout layer with rate 0.2. 
A subsequent $1 \times 1$ convolution outputs two logits, which are normalized with a channel-wise softmax to produce spatially varying weights $(w_{\mathrm{g}}, w_{\mathrm{r}})$. 
The fused features are then computed as
\begin{equation}
\begin{aligned}
\mathbf{F}_{\mathrm{fuse}}^{\mathrm{ici}} &= w_{\mathrm{g}} \odot \mathbf{F}_{\mathrm{g}}^{\mathrm{ici}} + w_{\mathrm{r}} \odot \mathbf{F}_{\mathrm{ici}}, \\
\mathbf{F}_{\mathrm{fuse}}^{\mathrm{ssi}} &= w_{\mathrm{g}} \odot \mathbf{F}_{\mathrm{g}}^{\mathrm{ssi}} + w_{\mathrm{r}} \odot \mathbf{F}_{\mathrm{ssi}}.
\end{aligned}
\label{eq:fuse}
\end{equation}
Finally, a $1 \times 1$ convolution with ReLU, serving as a contact adapter~$\mathbf{A}_{\mathrm{contact}}$, transforms the fused representations into the contact features
\begin{equation}
\begin{aligned}
\mathbf{F}_{\mathrm{c}}^{\mathrm{ici}} &= \mathbf{A}_{\mathrm{contact}}(\mathbf{F}_{\mathrm{fuse}}^{\mathrm{ici}}), \\
\mathbf{F}_{\mathrm{c}}^{\mathrm{ssi}} &= \mathbf{A}_{\mathrm{contact}}(\mathbf{F}_{\mathrm{fuse}}^{\mathrm{ssi}}).
\end{aligned}
\label{eq:contact}
\end{equation}
This attention-based fusion allows to adaptively combine information from the ground and randomized features when constructing the contact representations~$\mathbf{F}_{\mathrm{c}}^{\mathrm{ici}}$ and~$\mathbf{F}_{\mathrm{c}}^{\mathrm{ssi}}$.

\subsection{Foot contact decoder}
\label{sec:foot_contact_dec}
We follow HACO~\cite{jung2025learning} and adopt a Transformer-based network, that receives a contact token together with the image feature to predict dense contact, as foot contact decoder. 
The decoder processes each contact feature~$\mathbf{F}_{\mathrm{c}}^{\mathrm{ici}}$ and~$\mathbf{F}_{\mathrm{c}}^{\mathrm{ssi}}$ through consecutive self-attention and cross-attention and outputs dense foot contact logits after added with learnable contact initialization.
Formally, the contact decoder~$\mathrm{Dec}_{\mathrm{c}}$ maps the fused contact features to vertex-level logits,
\begin{equation}
\mathbf{C}^{\mathrm{ici}} = \mathrm{Dec}_{\mathrm{c}}(\mathbf{F}_{\mathrm{c}}^{\mathrm{ici}}), 
\qquad
\mathbf{C}^{\mathrm{ssi}} = \mathrm{Dec}_{\mathrm{c}}(\mathbf{F}_{\mathrm{c}}^{\mathrm{ssi}}),
\end{equation}
where $\mathbf{C}^{\mathrm{ici}}, \mathbf{C}^{\mathrm{ssi}} \in \mathbb{R}^{V}$ and $V=265$ is the number of foot vertices. 
The logits are then passed through a sigmoid function to obtain contact probabilities,
\begin{equation}
\hat{\mathbf{C}}^{\mathrm{ici}} = \sigma(\mathbf{C}^{\mathrm{ici}}), 
\qquad
\hat{\mathbf{C}}^{\mathrm{ssi}} = \sigma(\mathbf{C}^{\mathrm{ssi}}).
\end{equation}
Compared to HACO, which predicts contact for the MANO hand mesh with $778$ vertices, our decoder adapts the output dimension to the foot mesh with $265$ vertices and employs foot-specific regressors for supervision.

\begin{figure*}[t]
\begin{center}
\includegraphics[width=0.85\linewidth]{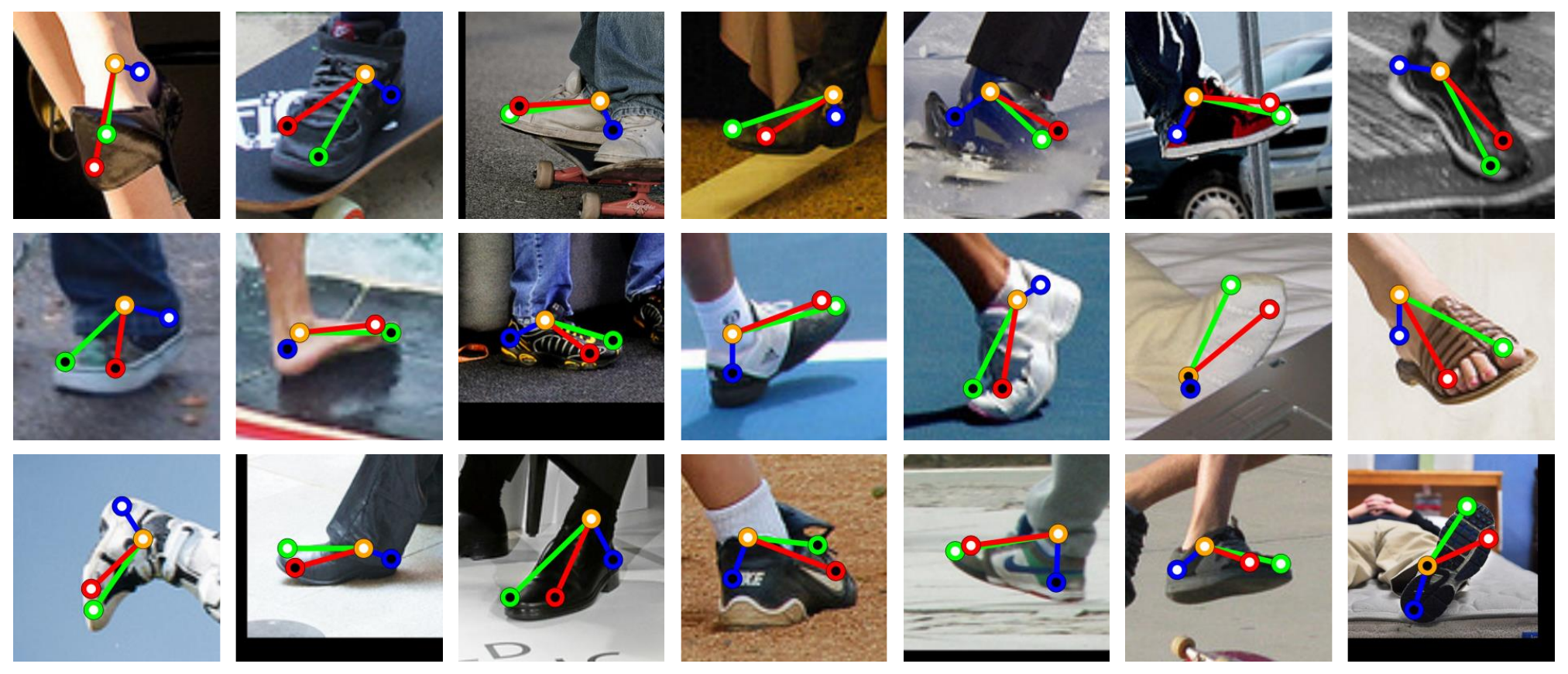}
\end{center}
\vspace{-4mm}
\caption{
\textbf{COFE Dataset.} We manually annotate joint-level foot contact for samples in OpenPose~\cite{cao2019openpose}, InstaVariety~\cite{kanazawa2019learning}, PennAction~\cite{zhang2013actemes}, and MPII~\cite{andriluka20142d} datasets. In the visualization, black indicates contacting joints and white represents non-contacting joints.}
\label{fig:openpose_dataset}
\vspace{-0.3cm}
\end{figure*}

\subsection{Final outputs and loss functions}
\label{sec:outputs_and_loss}

\noindent\textbf{Multi-level outputs.}
From the decoder we obtain vertex-level logits for both branches, $\mathbf{C}^{\mathrm{ici}}, \mathbf{C}^{\mathrm{ssi}} \in \mathbb{R}^{V}$, where $V=265$ refers to number of foot vertices. 
Following HACO~\cite{jung2025learning}, these logits are projected to coarser levels using regressors $\{\mathcal{J}_i^{v_i}\}_{i=1}^{N}$,
\begin{equation}
\begin{aligned}
\mathbf{C}_i^{\mathrm{ici}} &= \mathcal{J}_i^{v_i}\,\mathbf{C}^{\mathrm{ici}}, \quad i = 1, \ldots, N, \\
\mathbf{C}_i^{\mathrm{ssi}} &= \mathcal{J}_i^{v_i}\,\mathbf{C}^{\mathrm{ssi}}, \quad i = 1, \ldots, N.
\end{aligned}
\label{eq:multi_level_joint_regressor}
\end{equation}
where $N=3$ and $v_i \in \{V, 11, 3\}$. 
To obtain the contact probabilities, we apply a sigmoid after projection: 
\begin{equation}
\hat{\mathbf{C}}_i^{\mathrm{ici}} \;=\; \sigma(\mathbf{C}_i^{\mathrm{ici}}), 
\qquad
\hat{\mathbf{C}}_i^{\mathrm{ssi}} \;=\; \sigma(\mathbf{C}_i^{\mathrm{ssi}}).
\end{equation}
For the full mesh ($v_1=V$), this corresponds to the foot region of SMPL-X~\cite{pavlakos2019expressive} human body mesh. 
For $v_2=11$, we partition the foot into notable regions (five toes, heel, front, bottom, left, right, back) and use region means to define keypoints. 
For $v_3=3$, we follow OpenPose Human Foot Keypoints~\cite{cao2019openpose}. 
As there exists no official regressor that projects SMPL-X to OpenPose foot keypoints, we map big toe, small toe, and heel from our 11-joint definition to build SMPL-X to OpenPose regressor, which enables direct supervision with our COFE dataset introduced in Section~\ref{sec:openpose_foot_contact_dataset}.

\noindent\textbf{Loss functions.}
We train FECO end-to-end with the following total loss objective:
\begin{equation}
\mathcal{L} = \mathcal{L}_{\text{main}} + \mathcal{L}_{\text{style}} + \mathcal{L}_{\text{style-adv}} + \mathcal{L}_{\text{mask}} + \mathcal{L}_{\text{ground}}.
\end{equation}
The main loss $\mathcal{L}_{\text{main}}$ is a binary cross-entropy loss applied to the multi-level predictions of the main branch. 
The style loss $\mathcal{L}_{\text{style}}$ is the same loss applied to the style branch, but its gradients are restricted to the contact decoder in style branch. 
The style adversarial loss $\mathcal{L}_{\text{style-adv}}$ follows SagNets~\cite{nam2021reducing} to compute the binary cross-entropy loss between contact prediction from style branch and uniform distribution, where only the adversarial adapters $\mathbf{A}_{\mathrm{prev}}$ and $\mathbf{A}_{\mathrm{after}}$ are trained. 
The mask loss $\mathcal{L}_{\text{mask}}$ is the average of binary cross-entropy and Dice losses~\cite{milletari2016v} computed on the predicted foot segmentation. 
The ground loss $\mathcal{L}_{\text{ground}}$ is the sum of a pixel-height loss and a ground-normal loss,
\begin{equation}
\mathcal{L}_{\text{ground}} = \mathcal{L}_{\text{pixel-height}} + \mathcal{L}_{\text{ground-normal}},
\end{equation}
where $\mathcal{L}_{\text{pixel-height}}$ is mean absolute error on the scaled pixel height map~\cite{sheng2023pixht} and $\mathcal{L}_{\text{ground-normal}}$ is cosine similarity loss on ground normals. 
All losses are computed for FECO outputs from the clean image and two ProRandConv~\cite{choi2023progressive} augmentations, and then averaged. 
The ground loss~$\mathcal{L}_{\text{ground}}$ is applied symmetrically to both the main and style branches.
\section{Dataset: COFE}
\label{sec:openpose_foot_contact_dataset}

Despite the availability of accurate contact data from large-scale 3D datasets (Table~\ref{tab:dataset}), they lack in-the-wild data for training and evaluation.
To fill this gap, we introduce the COFE dataset, where we manually annotate binary foot contact labels on human foot keypoints from multiple 2D keypoint datasets: OpenPose~\cite{cao2019openpose}, InstaVariety~\cite{kanazawa2019learning}, PennAction~\cite{zhang2013actemes}, and MPII~\cite{andriluka20142d}.
After removing images with fewer than two valid foot joints, typically due to occlusion or annotation ambiguity, the dataset comprises 31,598 training and 1,567 testing images.
Representative samples are shown in Figure~\ref{fig:openpose_dataset}.
Since foot contact can be ambiguous in cases such as snowboarding, we define consistent annotation rules: snowboards are labeled as contact, skis as non-contact, and clothing as non-contact.
This avoids confusion with skateboards, which are never fixed to the feet.
Skis are only annotated as contact when they touch the ground or another surface, while foot–clothing contact (e.g., long dresses) is excluded due to its limited relevance to foot–ground or foot–object interaction.
We decided to exclude OpenPose for test set as our examination showed that it predominantly contains fully contacted feet; if all joints are manually predicted as contact, the precision, recall, and F1-score are 0.712, 0.858, and 0.760, respectively.
Hence, we construct our test set using the InstaVariety~\cite{kanazawa2019learning} dataset, which has diverse contact distribution and has video sequences that allow comparison with existing methods.

\begin{figure*}[t]
\begin{center}
\includegraphics[width=0.85\linewidth]{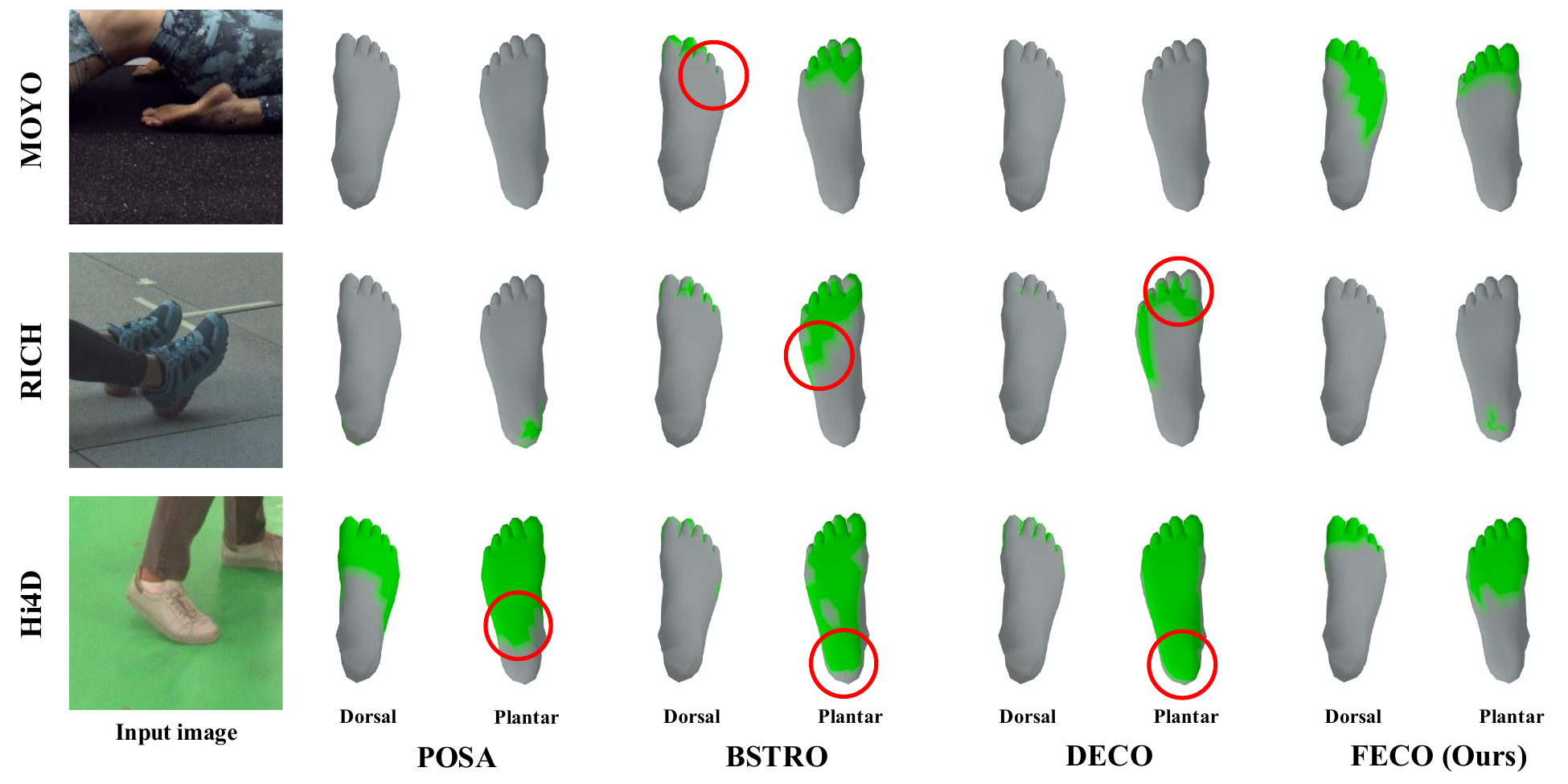}
\end{center}
\vspace{-7mm}
\caption{\textbf{Qualitative comparison of dense foot contact estimation with POSA~\cite{hassan2021populating}, BSTRO~\cite{huang2022capturing}, and DECO~\cite{tripathi2023deco} on MOYO~\cite{tripathi20233d}, RICH~\cite{huang2022capturing}, Hi4D~\cite{yin2023hi4d} dataset.} Red circles indicate exemplar regions that FECO outperforms previous methods.}
\label{fig:sota_qual_dense_foot_contact}
\vspace{-0.3cm}
\end{figure*}
\section{Implementation details}
PyTorch~\cite{paszkepytorch} is used for implementation.
Our backbone is initialized publicly released weights of ViT-Huge pretrained with ImageNet~\cite{deng2009imagenet}.
We apply data augmentations such as random scaling, cropping, and rotation.
To enhance robustness against degraded inputs, we further introduce low-resolution, noise, and blur perturbations during the augmentation.
We employ the AdamW optimizer~\cite{loshchilov2018decoupled} with a learning rate of $10^{-5}$ and a mini-batch size of 4.
To ensure stable convergence, the learning rate is decayed by a factor of 0.9 after the 5th and 10th epochs.
We train FECO for 10 epochs on a single NVIDIA A6000 GPU.
\section{Experiments}

\begin{table}[t]
\centering
\footnotesize
\setlength{\tabcolsep}{2.5pt}
\renewcommand{\arraystretch}{0.9}
\caption{\textbf{Ablation of low-level randomization on MMVP~\cite{zhang2024mmvp}.}}
\vspace{-0.5em}
\begin{tabular}{@{}lccc@{}}
\toprule
Methods & Precision $\uparrow$ & Recall $\uparrow$ & F1-Score $\uparrow$ \\
\midrule
w/o Low-Level Rand. & 0.544 & 0.584 & 0.555 \\
w/ Low-Level Rand. (Ours) & \textbf{0.563} & \textbf{0.613} & \textbf{0.577} \\
\bottomrule
\end{tabular}
\label{tab:abl_low_level}
\vspace{-0.2cm}
\end{table}

\begin{table}[t]
\centering
\footnotesize
\setlength{\tabcolsep}{3pt}
\caption{\textbf{Ablation of shoe randomization on MMVP~\cite{zhang2024mmvp}.}}
\vspace{-0.5em}
\begin{tabular}{@{}ccccc@{}}
\toprule
Content Rand. & Style Rand. & Precision $\uparrow$ & Recall $\uparrow$ & F1-Score $\uparrow$ \\ 
\midrule
\ding{55} & \ding{55} & 0.515 & 0.610 & 0.522 \\
\ding{51} & \ding{55} & 0.504 & \textbf{0.651} & 0.531 \\
\ding{55} & \ding{51} & 0.541 & 0.595 & 0.554 \\
\ding{51} & \ding{51} & \textbf{0.563} & 0.613 & \textbf{0.577} \\ 
\bottomrule
\end{tabular}
\label{tab:abl_cont_sty_rand}
\end{table}

\subsection{Datasets}
We choose 10 datasets with diverse foot interactions with PROX~\cite{hassan2019resolving}, EgoBody~\cite{zhang2022egobody}, RICH~\cite{huang2022capturing} for foot-scene interaction, BEHAVE~\cite{bhatnagar2022behave, xie2024rhobin}, InterCap~\cite{huang2022intercap} for foot-object interaction, MOYO~\cite{tripathi20233d}, MMVP~\cite{zhang2024mmvp}, MotionPRO~\cite{ren2025motionpro} for foot-ground interaction, and Hi4D~\cite{yin2023hi4d} for foot-body interaction.
Additionally, our proposed COFE dataset, introduced in Section~\ref{sec:openpose_foot_contact_dataset}, is utilized for in-the-wild training samples.
To reduce redundancy and balance the ratio between datasets, we sample 2, 5, 5, 10, 50, 30 for BEHAVE, EgoBody, Hi4D, InterCap, MotionPRO, MOYO dataset.
Our main evaluation dataset is MMVP.
For shoe content randomization, we use UT Zappos50K dataset~\cite{yu2014fine}, which contains 50,025 images across four major categories: shoes, sandals, slippers, and boots, as well as four relative attributes at instance level: open, pointy, sporty, and comfort.

\subsection{Evaluation metrics}
To evaluate dense foot contact estimation, we report precision, recall, and F1 score. 
Given predicted per-vertex probabilities $\hat{\mathbf{C}}\in[0,1]^V$ and binary ground truth $\mathbf{C_\mathrm{GT}}\in\{0,1\}^V$, we threshold $\hat{\mathbf{C}}$ to obtain binary predictions and compute per-sample precision, recall, and F1, then average over the dataset. 
Samples without any positive contact are excluded as recall and F1 are not well defined in such cases.

\begin{table}[t]
\centering
\footnotesize
\setlength{\tabcolsep}{2.7pt}
\caption{\textbf{Ablation of ground-aware learning on MMVP~\cite{zhang2024mmvp}.}}
\vspace{-0.5em}
\begin{tabular}{@{}cccccc@{}}
\toprule
Ground norm. & PH map & Spatial attn. & Precision $\uparrow$ & Recall $\uparrow$ & F1-Score $\uparrow$ \\
\midrule
\ding{55} & \ding{55} & \ding{55} & 0.482 & 0.527 & 0.506 \\
\ding{51} & \ding{55} & \ding{55} & 0.518 & 0.560 & 0.527 \\
\ding{51} & \ding{51} & \ding{55} & 0.554 & 0.610 & 0.569 \\
\ding{51} & \ding{51} & \ding{51} & \textbf{0.563} & \textbf{0.613} & \textbf{0.577} \\
\bottomrule
\end{tabular}
\label{tab:abl_ground_aware}
\end{table}

\begin{table}[t]
\centering
\footnotesize
\setlength{\tabcolsep}{3.5pt}
\caption{\textbf{Ablation of training dataset on COFE dataset.}}
\vspace{-0.5em}
\begin{tabular}{@{}ccccc@{}}
\toprule
3D Mocap datasets & COFE dataset & Precision $\uparrow$ & Recall $\uparrow$ & F1-Score $\uparrow$ \\ 
\midrule
\ding{51} & \ding{55} & 0.494 & 0.464 & 0.450 \\
\ding{51} & \ding{51} & \textbf{0.553} & \textbf{0.516} & \textbf{0.515} \\
\bottomrule
\end{tabular}
\label{tab:abl_openpose}
\end{table}

\subsection{Ablation study}
\noindent\textbf{Effectiveness of low-level randomization.}
Table~\ref{tab:abl_low_level} shows that progressive low-level style randomization improves performance on MMVP~\cite{zhang2024mmvp}. 
Compared to the variant without low-level randomization, precision increases by 3.5\%, recall by 5.0\%, and F1-score by 4.0\%. 
This indicates that our low-level randomization effectively helps FECO rely less on spurious low-level appearance cues and generalize better across various shoe styles.

\noindent\textbf{Effectiveness of style-content randomization.}
Table~\ref{tab:abl_cont_sty_rand} shows that content randomization alone yields the highest recall but lowers precision, suggesting stronger coverage at the cost of false positives. 
Style randomization alone increases precision while maintaining a competitive recall, indicating improved robustness to style shifts. 
Combining both content and style randomization produces the best trade-off with highest F1-score, confirming that disentangling content and style and exposing FECO to diverse shoe appearances from shoe image dataset UT Zappos50K dataset~\cite{yu2014fine} yields complementary gains.

\noindent\textbf{Effectiveness of ground-aware learning.}
Table~\ref{tab:abl_ground_aware} evaluates ground-aware learning. 
Adding ground normals to the baseline improves F1-score, showing that global orientation of the support surface is informative. 
Introducing pixel height maps further lifts the F1-score by providing dense geometric context that correlates with the ground. 
Finally, spatial attention between randomized features and ground features yields the best result, indicating that adaptive fusion of shoe style-invariant learning and ground-aware learning leads to harmonious gains across all metrics.

\begin{table}[t]
\centering
\footnotesize
\setlength{\tabcolsep}{4pt}
\caption{\textbf{Comparison of feature-level style randomization techniques on MMVP~\cite{zhang2024mmvp}.} 
† denotes re-implemented results.}
\vspace{-0.5em}
\begin{tabular}{@{}lccc@{}}
\toprule
Method & Precision $\uparrow$ & Recall $\uparrow$ & F1-Score $\uparrow$ \\ 
\midrule
BIN~\cite{nam2018batch} & 0.505 & 0.351 & 0.396 \\
MixStyle~\cite{zhoudomain} & 0.437 & 0.463 & 0.448 \\
SagNets~\cite{nam2021reducing} & 0.451 & 0.564 & 0.511 \\
†LatentDR~\cite{liu2024latentdr} & 0.534 & 0.574 & 0.542 \\
\midrule
\textbf{Shoe Style–Content Rand. (Ours)} & \textbf{0.563} & \textbf{0.613} & \textbf{0.577} \\
\bottomrule
\end{tabular}
\label{tab:sota_feat_sty_rand}
\end{table}

\begin{table}[t]
\centering
\footnotesize
\setlength{\tabcolsep}{2.8pt}
\renewcommand{\arraystretch}{0.9}
\caption{\textbf{Comparison with SOTA methods of dense foot contact estimation on MMVP~\cite{zhang2024mmvp}.}}
\vspace{-0.5em}
\begin{tabular}{@{}lccc@{}}
\toprule
Methods & Precision $\uparrow$ & Recall $\uparrow$ & F1-Score $\uparrow$ \\ 
\midrule
POSA~\cite{hassan2021populating} & 0.276 & 0.308 & 0.255 \\
BSTRO~\cite{huang2022capturing} & 0.436 & 0.538 & 0.464 \\
DECO~\cite{tripathi2023deco} & 0.374 & 0.511 & 0.409 \\
\midrule
FECO~(Ours) & \textbf{0.563} & \textbf{0.613} & \textbf{0.577} \\
\bottomrule
\end{tabular}
\label{tab:sota_foot_vert_cont}
\vspace{-0.2cm}
\end{table}

\noindent\textbf{Effectiveness of COFE dataset.}
Table~\ref{tab:abl_openpose} analyzes the impact of adding our proposed COFE dataset as additional training data. 
Compared with FECO trained with only 3D motion capture~(mocap) datasets, FECO trained additionally with our COFE dataset is better in all metrics of precision, recall, and F1-score. 
The gains suggest that COFE supplies diverse in-the-wild appearances and various foot interactions that complement 3D mocap datasets and reduce overfitting to appearances from the 3D mocap datasets.

\subsection{Comparison with state-of-the-art methods}
\noindent\textbf{Dense foot contact estimation.}
Table~\ref{tab:sota_foot_vert_cont} demonstrates strong performance of FECO on MMVP dataset~\cite{zhang2024mmvp} compared to prior methods by a clear margin. 
Compared to POSA~\cite{hassan2021populating}, BSTRO~\cite{huang2022capturing}, and DECO~\cite{tripathi2023deco}, FECO achieves the highest precision, recall, and F1-score, demonstrating strong performance across both positive and negative predictions. 
Such gains are largely attributed to diversity of training data, shoe style–invariant learning that mitigates appearance bias and ground-aware learning that injects explicit support-surface geometry.

\noindent\textbf{Joint-level foot contact estimation.}
In Table~\ref{tab:sota_foot_joint_cont}, we compare joint-level foot contact estimation methods on our COFE dataset.
Most existing methods rely on zero-velocity heuristics or motion cues, making them unsuitable for single-image inputs.
Hence, for fair comparison, we construct the test split with only video sequences.
Evaluation is conducted on two joints, toe and heel, following the common definition in the prior works.
To evaluate, we use heel contact directly and aggregate the two toe contacts with a logical OR operation.
It is worth noting that FECO is the only method that does not have access to temporal information.
Even under these settings, FECO achieves the best precision, recall, and F1-score, showing that dense contact reasoning from single images generalizes well to joint-level estimation, surpassing methods that exploit temporal cues.

\begin{table}[t]
\centering
\footnotesize
\setlength{\tabcolsep}{3pt}
\renewcommand{\arraystretch}{0.9}
\caption{\textbf{Comparison with SOTA methods of joint-level foot contact estimation on COFE dataset.}}
\vspace{-0.5em}
\begin{tabular}{@{}lccc@{}}
\toprule
Methods & Precision $\uparrow$ & Recall $\uparrow$ & F1-Score $\uparrow$ \\ 
\midrule
Footskate Reducer~\cite{zou2020reducing} & 0.399 & 0.271 & 0.301 \\
WHAM~\cite{shin2024wham} & 0.347 & 0.431 & 0.363 \\
\midrule
FECO~(Ours) & \textbf{0.553} & \textbf{0.516} & \textbf{0.515} \\
\bottomrule
\end{tabular}
\label{tab:sota_foot_joint_cont}
\vspace{-0.2cm}
\end{table}
\section{Conclusion}
We propose FECO, a robust and generalizable framework for dense foot contact estimation that incorporates shoe style–invariant and ground-aware learning. 
To achieve style invariance, we introduce shoe style–content randomization using an external shoe image dataset. 
For ground-aware learning, we supervise both pixel height maps and ground normals to capture fine-grained and global foot–ground interactions. 
With the proposed COFE dataset, our FECO surpasses previous state-of-the-art methods by a significant margin and demonstrates robustness and generalization.

\noindent\textbf{Acknowledgements.}
This work was supported in part by the IITP grants [No. RS-2021-II211343, Artificial Intelligence Graduate School Program (Seoul National University), No. RS-2025-02303870, No.2022-0-00156] funded by the Korea government (MSIT).

\clearpage

\maketitlesupplementary

\appendix

\setcounter{section}{0}
\setcounter{table}{0}
\setcounter{figure}{0}

\renewcommand{\thesection}{S\arabic{section}}
\renewcommand{\thetable}{S\arabic{table}}   
\renewcommand{\thefigure}{S\arabic{figure}}

\renewcommand{\theHsection}{S\arabic{section}}
\renewcommand{\theHtable}{S\arabic{table}}
\renewcommand{\theHfigure}{S\arabic{figure}}

In this supplementary material, we provide additional technical details and experimental results that were omitted from the main manuscript due to space constraints.
The contents are summarized below:
\begin{itemize}
\item \ref{sec:supp_cofe_configuration}. Configuration of COFE dataset
\item \ref{sec:supp_training_loss_detail}. Details of training FECO
\item \ref{sec:supp_joint_def_reg}. Details of joint definitions
\item \ref{sec:supp_dense_foot_label}. Details of dense foot contact labels
\item \ref{sec:supp_vis_ground_aware}. Visualization of ground-aware learning
\item \ref{sec:supp_quant_more_datasets}. Quantitative results on more datasets
\item \ref{sec:supp_diff_backbones}. Quantitative results on different backbones
\item \ref{sec:supp_computation}. Computational requirements
\item \ref{sec:supp_more_qual}. More qualitative results
\item \ref{sec:limit_and_impacts}. Limitations and societal impacts
\end{itemize}

\section{Configuration of COFE dataset}
\label{sec:supp_cofe_configuration}
\begin{figure}[t]
  \centering
\includegraphics[width=\linewidth,keepaspectratio]{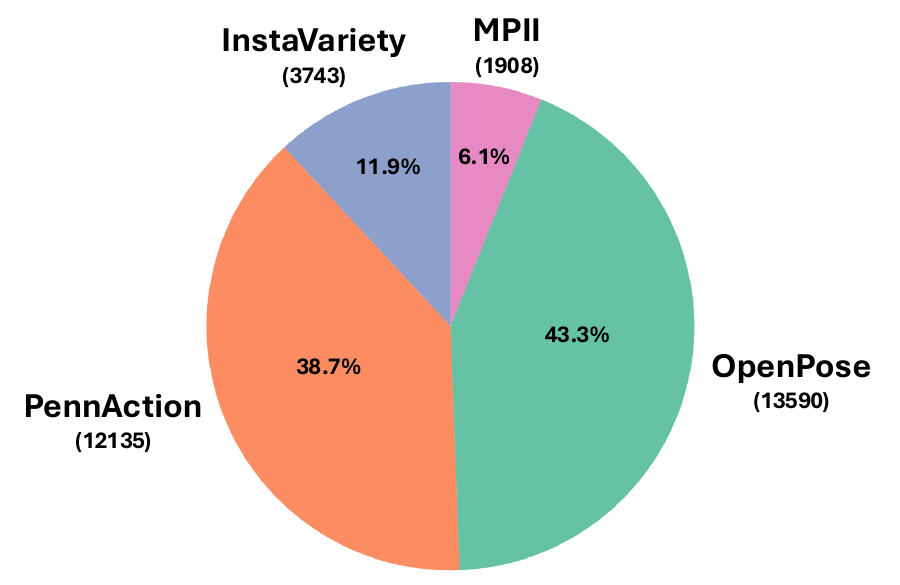}
  \vspace{-4mm}
  \caption{\textbf{COFE dataset statistics.}
  We visualize the dataset configuration of our proposed COFE dataset, which consists of foot image samples in OpenPose~\cite{cao2019openpose}, InstaVariety~\cite{kanazawa2019learning}, PennAction~\cite{zhang2013actemes}, and MPII~\cite{andriluka20142d}. We only include training samples.}
  \label{fig:cofe_piechart}
\end{figure}

We construct the COFE dataset by aggregating foot image samples from OpenPose~\cite{cao2019openpose}, InstaVariety~\cite{kanazawa2019learning}, PennAction~\cite{zhang2013actemes}, and MPII~\cite{andriluka20142d}. 
Table~\ref{tab:cofe_split_statistics} presents the train--test split of the aggregated dataset, and Figure~\ref{fig:cofe_piechart} illustrates the relative proportion of training samples from each source. 
Regarding the training set, COFE dataset contains 43.3\% samples from OpenPose, 38.7\% from PennAction, 11.9\% from InstaVariety, and 6.1\% from MPII.

For image-based datasets of OpenPose and MPII, we only include frames in which the feet are clearly visible. 
For video-based datasets of PennAction and InstaVariety, we additionally filter out static clips, particularly those of which the person remains standing upright throughout the sequence, as fully contacted foot states are already sufficiently covered in image datasets. 
To ensure annotation quality, all samples are manually labeled following the OpenPose foot keypoint definition (big toe, small toe, and heel). 
Moreover, we primarily use videos containing a single person, so as to avoid erroneous keypoint detections caused by multiple individuals within a frame. 
Notably, the MPII videos are generally very short (less than three seconds), which limits their utility as test set of COFE dataset.
Hence, we primarily use MPII dataset only for training.

Although OpenPose includes a small set of test samples, we found that nearly all of them correspond to fully contacted feet. 
As a result, the evaluation is biased: for example, when predicting uniformly full contact across the entire OpenPose test set, the resulting precision, recall, and F1-score are already 0.712, 0.858, and 0.760, respectively. 
This suggests that OpenPose is not suitable for robust evaluation. 
Also, since previous methods for foot contact estimation are video-based and temporal in nature, we choose to perform evaluation solely on the InstaVariety test split samples within the COFE dataset. 
This choice allows us to coherently evaluate joint-level foot contact estimation.

\begin{table}[t]
\centering
\small
\setlength{\tabcolsep}{4pt}
\caption{\textbf{Data split for COFE dataset.}}
\begin{tabular}{lccc}
\toprule
Dataset & Image / Video & Train  & Test \\
\midrule
OpenPose~\cite{cao2019openpose} & Image & 13,590 & 464 \\
PennAction~\cite{zhang2013actemes} & Video & 12,135 & - \\
InstaVariety~\cite{kanazawa2019learning} & Video & 3,743  & 1,103 \\
MPII~\cite{andriluka20142d} & Image & 1,908  & - \\
\bottomrule
\end{tabular}
\label{tab:cofe_split_statistics}
\end{table}

\begin{figure*}[t]
\begin{center}
\includegraphics[width=1.0\linewidth]{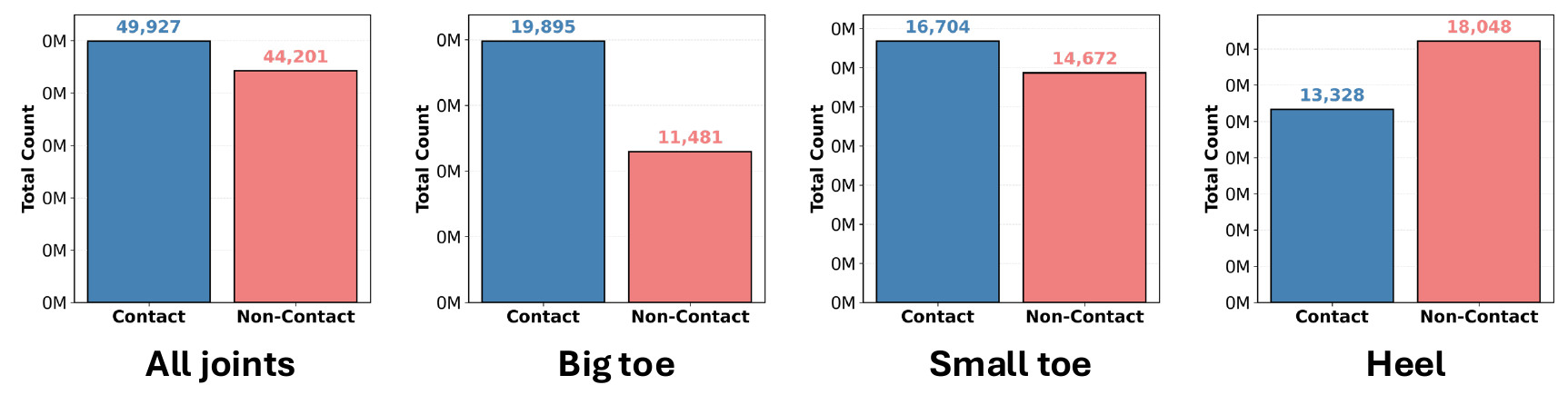}
\end{center}
\vspace{-3.5mm}
\caption{
\textbf{Contact and non-contact distribution of COFE dataset.}}
\label{fig:supp_cofe_joint_stat}
\vspace{-0.3cm}
\end{figure*}

As shown in Figure~\ref{fig:supp_cofe_joint_stat}, the COFE training set maintains a relatively balanced distribution between contacting and non-contacting joints, with 49,927 contacting and 44,201 non-contacting cases overall. 
At the joint level, big toe contacts account for 63.4\% of its annotations, small toe for 53.3\%, while heel contacts are slightly underrepresented at 42.5\%. 
This indicates that, the heel joint is slightly less frequent in contact compared to the toes, but the dataset still provides sufficient coverage across all three joints. 

\begin{figure*}[htbp]
\begin{center}
\includegraphics[width=1.0\linewidth]{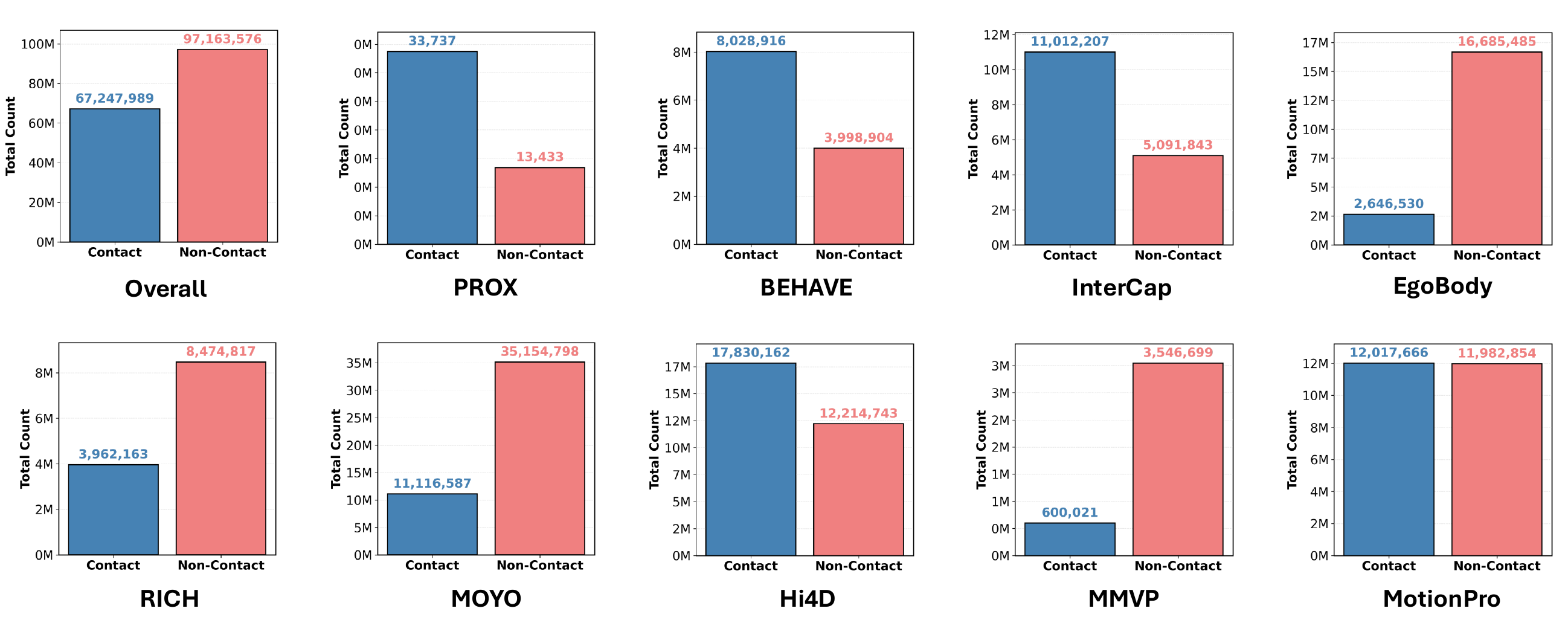}
\end{center}
\vspace{-3.5mm}
\caption{
\textbf{Contact and non-contact distribution of training datasets for FECO.}}
\label{fig:supp_contact_vs_noncontact}
\vspace{-0.3cm}
\end{figure*}
\begin{table}[htbp]
\centering
\caption{\textbf{Ground plane configuration.} Negative height refers to datasets whose coordinate system is defined such that smaller values along the height axis correspond to higher positions.}
\vspace{0.6em}
\scalebox{0.87}{\begin{tabular}{lccccc} \toprule
Dataset & Height axis & Ground axis & Negative height \\
\midrule
PROX~\cite{hassan2019resolving} & z-axis & x-axis, y-axis & \ding{55} \\ % Checked mesh
BEHAVE~\cite{bhatnagar2022behave}  & y-axis & x-axis, z-axis & \ding{51} \\ % Checked mesh
InterCap~\cite{huang2022intercap} & y-axis & x-axis, z-axis & \ding{51} \\ % Checked mesh
EgoBody~\cite{zhang2022egobody} & y-axis & x-axis, z-axis & \ding{55} \\ % Checked mesh
RICH~\cite{huang2022capturing} & y-axis & x-axis, z-axis & \ding{51} \\ % Checked mesh
MOYO~\cite{tripathi20233d} & z-axis & x-axis, y-axis & \ding{55} \\ % Checked mesh
Hi4D~\cite{yin2023hi4d} & y-axis & x-axis, z-axis & \ding{55} \\ % Checked mesh
MMVP~\cite{zhang2024mmvp} & y-axis & x-axis, z-axis & \ding{51} \\ % Checked mesh
MotionPRO~\cite{ren2025motionpro} & y-axis & x-axis, z-axis & \ding{55} \\

\bottomrule
\end{tabular}}
\vspace{-0.3cm}
\label{tab:ground_mesh}
\end{table}

\begin{figure*}[t]
\begin{center}
\includegraphics[width=0.85\linewidth]{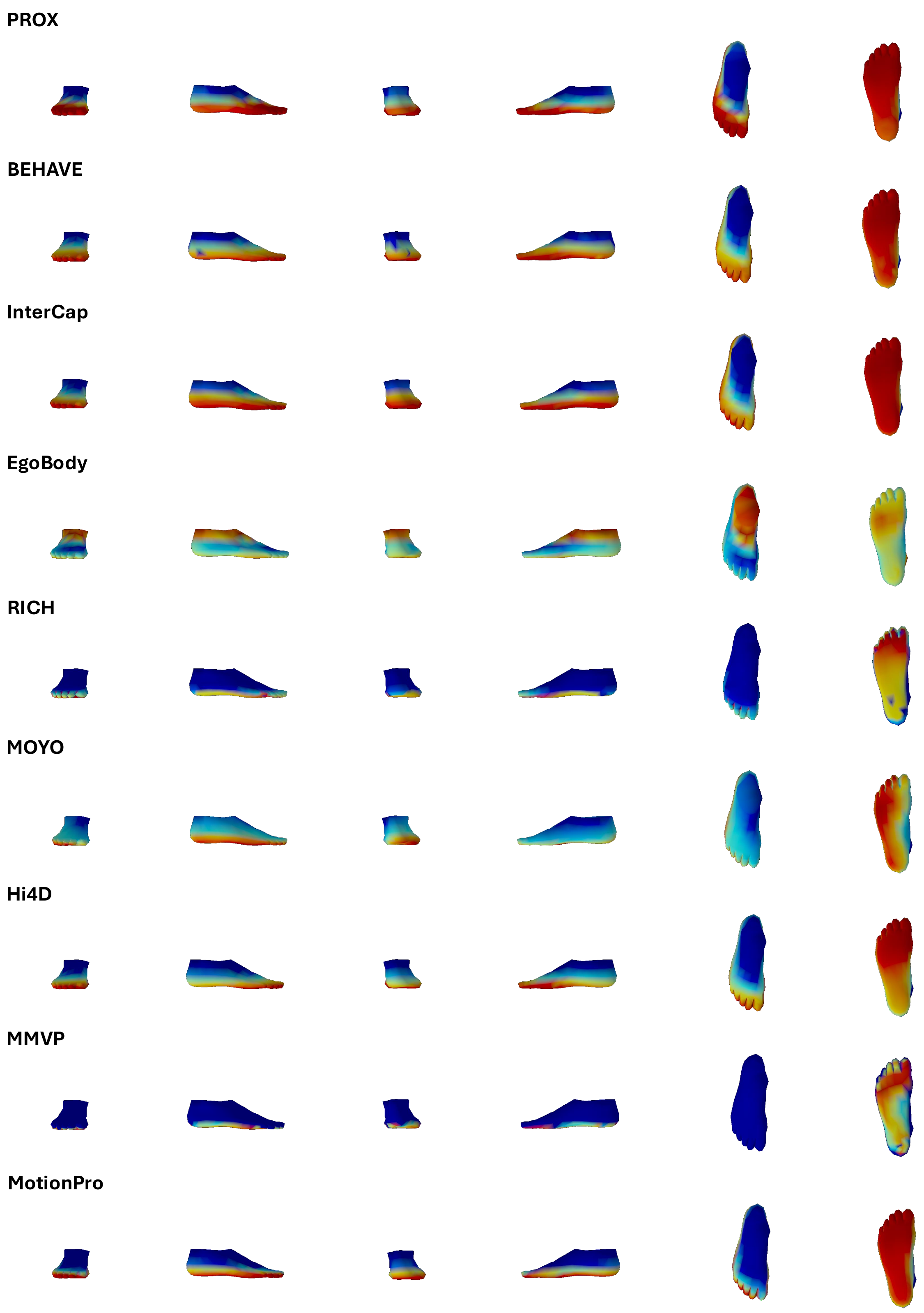}
\end{center}
\vspace{-3.5mm}
\caption{
\textbf{Dataset-wise dense foot contact mean.} These heatmaps show mean foot contact of ground-truth contacts from PROX~\cite{hassan2019resolving}, BEHAVE~\cite{bhatnagar2022behave}, InterCap~\cite{huang2022intercap}, EgoBody~\cite{zhang2022egobody}, RICH~\cite{huang2022capturing}, MOYO~\cite{tripathi20233d}, Hi4D~\cite{yin2023hi4d}, MMVP~\cite{zhang2024mmvp}, MotionPro~\cite{ren2025motionpro} dataset.}
\label{fig:supp_contact_heatmap}
\vspace{-0.3cm}
\end{figure*}

When compared with the overall distribution from other training datasets used for FECO (excluding COFE), a clearer contrast emerges. 
As visualized in Figure~\ref{fig:supp_contact_vs_noncontact}, the combined datasets exhibit a skew toward non-contact, with roughly 41\% contacting versus 59\% non-contacting joints in total. 
Moreover, these datasets show contacts concentrated at the big and small toes, with the heel being significantly underrepresented as in Figure~\ref{fig:supp_contact_heatmap}. 
By contrast, COFE not only achieves a closer balance between contact and non-contact, but also distributes annotations more evenly across joints. 
This makes COFE a complementary resource that mitigates the biases present in existing datasets and provides a more stable training signal for joint-level foot contact estimation.

\begin{figure*}[htbp]
\begin{center}
\includegraphics[width=1.0\linewidth]{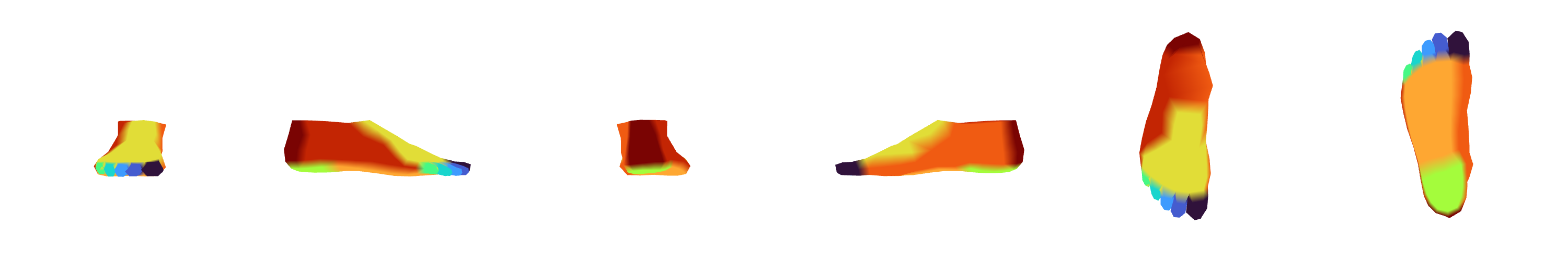}
\end{center}
\vspace{-3.5mm}
\caption{
\textbf{Foot part segmentation.} We build foot part segmentation that consists of 11 parts. Each part is defined by 5 toes (blue), heel (neon lime green), front (yellow), bottom (bright orange), left (dark orange), right (red), back (dark red).}
\label{fig:foot_part_seg}
\vspace{-0.3cm}
\end{figure*}

\section{Details of training FECO}
\label{sec:supp_training_loss_detail}

The original SagNets~\cite{nam2021reducing} adopt three separate optimizers, where one trains the content-biased network with a task loss and two others train the style-biased network with both a task loss and an adversarial loss. 
While this has a reasonable computational cost for image classification task, this strategy becomes computationally expensive when extended to dense foot contact estimation. 
To reduce the cost, we train our model in an end-to-end manner with a single optimizer, which allows the backbone network (\textit{e.g.,} ViT, ResNet), the foot contact decoders, the ground-aware decoders, and the style branch to be optimized jointly. 
This is achieved with gradient detaching and module freezing for irrelevant modules to resemble the training of the original SagNets. 
Moreover, all of the operations after low-level style randomization are conducted three parallel operation onto the original input and two low-level style randomized images with ProRandConv~\cite{choi2023progressive}.
This allows FECO to reduce low-level style bias, which improve model's stability and robustness on unseen shoe styles. 
Below, we further explain in detail on the process and clarify our design choice on training.

\noindent\textbf{Gradient detaching for foot segmentation.}
During the forward pass, the backbone produces multi-level intermediate features~$\{\mathbf{F}_{l}\}_{l=1}^{n-1}$ and a final high-resolution feature map~$\mathbf{F}_n$, where there are $n$ layers for the backbone network. 
From these features~$\{\mathbf{F}_{l}\}_{l=1}^{n}$, a foot segmentation mask~$\mathbf{M}_{\mathrm{f}}$ is predicted and binarized to detect the spatial region of the foot.
We intentionally detach the gradient flow from later modules that utilize the foot segmentation mask~$\mathbf{M}_{\mathrm{f}}$, so that the foot segmentation mask decoder is solely trained by the foot segmentation loss~$\mathcal{L}_{\text{mask}}$.

\noindent\textbf{Gradient detaching and module freezing for end-to-end training.}
There are two major components of training techniques of SagNets that requires modification to enable training of FECO on a single optimizer.
First, the style-biased loss of SagNets only train style-biased network., which each corresponds to style loss~$\mathcal{L}_{\text{style}}$ and foot contact decoder in our style branch.
This is implemented in SagNets by adopting a separate optimizer that only optimizes the parameters in style-biased network with a dedicated optimizer.
However, this can be easily implemented by simply detaching the gradient flow for the input of style-biased network.
For FECO, we therefore detach the graident flow of input content invariant feature towards foot contact decoder.
This allows our style loss~$\mathcal{L}_{\text{style}}$ to only train foot contact decoder.
However, the challenging part is that we need to obtain gradient from the same foot contact decoder in the style path of our FECO to allow adversarial training with our style adversarial loss~$\mathcal{L}_{\text{style-adv}}$.
In SagNets, the adversarial loss is applied to the affine parameters of the batch normalization layers within ResNet backbone.
Similar with their style loss, they also utilize dedicated optimizer that only optimizes the affine parameters with the adversarial loss.
Our FECO instead builds two adapters~$\mathbf{A}_{\mathrm{prev}}$ and $\mathbf{A}_{\mathrm{after}}$, which each adapts multi-level features~$\{\mathbf{F}_{l}\}_{l=1}^{n}$ and adapts features after content randomization.
Nevertheless, there is another challenge.
The adapters, which are our replacement of the affine parameters in ResNet backbone from SagNets, are at the initial stages of the FECO model while adversarial loss should only train the adapters~$\mathbf{A}_{\mathrm{prev}}$ and $\mathbf{A}_{\mathrm{after}}$ without influencing subsequent modules such as foot contact decoder in style branch of FECO.
This is difficult as loss inevitably influences all subsequent modules by nature.
To tackle this issue, we freeze all parameters within foot contact decoder within style branch of FECO and make only the adapters~$\mathbf{A}_{\mathrm{prev}}$ and $\mathbf{A}_{\mathrm{after}}$ to be trained with adversarial loss~$\mathcal{L}_{\text{style-adv}}$.
This enables us to strictly follow the training process of SagNets while utilizing only a single optimizer, which significantly decreases the computational burden of training.

\section{Details of joint definitions}
\label{sec:supp_joint_def_reg}
In Figure~\ref{fig:foot_part_seg}, we present the foot part segmentation used to extract foot joints for joint level foot contact supervision. 
This corresponds to $v_2=11$. 
The most widely used foot joint definition of OpenPose~\cite{cao2019openpose} provides a foot joint definition that corresponds to $v_3=3$, which is highly sparse and captures contact only at the big toe, small toe, and heel, limiting coverage of the remaining foot surface. 
But, we need a denser joint representation that covers the entire foot surface. 
Therefore, we start from the SMPL-X~\cite{pavlakos2019expressive} foot mesh, which is a subset of the full body model.
We partition the foot mesh into eleven intuitive parts that align with functional regions of the foot. 
First, we segment the five toes, visualized from dark to light blue in Figure~\ref{fig:foot_part_seg}. 
Then, we segment the heel to maintain compatibility with the OpenPose joint definition. 
Furthermore, we segment the remaining plantar surface that frequently contacts the ground. 
Finally, we divide the dorsal surface into four regions in the left, right, front, and back directions. 
This yields eleven parts used for both foot part segmentation and the associated joints. 
Each foot joint is then defined as the mean of the vertex coordinates within its corresponding part.

\begin{table*}[htbp]
\centering
\setlength{\tabcolsep}{4pt}
\caption{\textbf{Computational requirements of various backbone configurations.}}
\scalebox{0.87}{\begin{tabular}{lcccccc} \toprule
Model & Backbone & Train Memory~(MB) & Test Memory~(MB) & Params.~(M) & Speed~(fps) & GFLOPs \\
\midrule
FECO & ViT-H~\cite{dosovitskiy2020image} & 34,328 & 6,418 & 964.64 & 20.04 & 190.67 \\
FECO & ViT-L~\cite{dosovitskiy2020image} & 19,683 & 4,528 & 554.37 & 20.86 & 73.45 \\
FECO & ViT-B~\cite{dosovitskiy2020image} & 12,269 & 3,304 & 270.44 & 24.86 & 24.67 \\
FECO & ViT-S~\cite{dosovitskiy2020image} & 8,743 & 2,652 & 137.38 & 23.53 & 6.81 \\
\midrule
FECO & ResNet-152~\cite{he2016deep} & 24,410 & 3,680 & 335.66 & 17.98 & 99.75  \\
FECO & ResNet-101~\cite{he2016deep} & 21,732 & 3,644 & 320.02 & 20.82 & 87.49  \\
FECO & ResNet-50~\cite{he2016deep} & 19,037 & 3,570 & 301.03 & 24.16 & 72.60 \\
FECO & ResNet-34~\cite{he2016deep} & 4,757 & 2,420 & 104.79 & 41.07 & 4.11 \\
FECO & ResNet-18~\cite{he2016deep} & 4,463 & 2,378 & 94.68 & 42.98 & 2.26 \\

\bottomrule
\end{tabular}}
\vspace{-0.3cm}
\label{tab:supp_comp_require}
\end{table*}

\section{Details of dense foot contact labels}
\label{sec:supp_dense_foot_label}
Following the previous work on hand contact estimation~\cite{jung2025learning}, we implement distance-based thresholding with Trimesh library~\cite{trimesh} to gather ground-truth dense foot contact labels.
However, unlike HACO~\cite{jung2025learning}, that had access to mesh of the interacting entity~(\textit{i.e.,} 3D object mesh, 3D scene mesh), we only have a few datasets~\cite{hassan2019resolving, zhang2022egobody, huang2022capturing} that provide 3D scene mesh.
Therefore, in order to also leverage 3D motion capture datasets~\cite{bhatnagar2022behave, huang2022intercap, tripathi20233d, yin2023hi4d, zhang2024mmvp, ren2025motionpro} that do not provide 3D scene mesh, we extract 3D ground mesh and conduct distance-based thresholding between the 3D ground mesh and 3D foot mesh.

To extract ground mesh from datasets without 3D scene mesh, we fit a parametric ground plane to each capture sequence.
Specifically, we aggregate candidate ground points by collecting the vertices that are lowest in physical height (closest to the real-world ground surface) from both the body and interacting object meshes (only if provided) across frames, and then apply a regression-based plane fitting strategy to obtain coefficients $(a, b, c)$ of the plane $h = a g_1 + b g_2 + c$, where $h$ denotes the height axis (\textit{i.e.}, y-axis of the xyz coordinate system) and $g_1, g_2$ denote the ground axes (\textit{i.e.}, x-axis and z-axis of the xyz coordinate system).

We also extract ground meshes from datasets with 3D scene mesh.
To extract ground mesh from datasets with 3D scene mesh, we directly leverage the 3D scene mesh. 
We first sample the scene mesh vertices and compute their height values with respect to the vertical axis, taking into account the coordinate convention of each dataset. 
Among these vertices, we retain only those within the lowest $p$\% percentile in height, which effectively restricts candidate points to the near-ground region while discarding elevated structures and outliers. 
We then apply RANSAC-based plane fitting~\cite{fischler1981random} to these candidate points, repeating the procedure multiple times and selecting the plane with the widest spatial support, determined by how many scene mesh vertices fall close to the fitted plane within a distance threshold.
This yields the ground plane parameters $(a, b, c)$ of $h = a g_1 + b g_2 + c$, consistent with the coordinate definition used for datasets without 3D scene mesh.

Once the ground plane is estimated, we compute signed distances between the 3D foot mesh vertices and the plane, and assign contact labels when the magnitude of the distance falls within a dataset-specific tolerance. 
The tolerance values are determined by manually inspecting the dense foot contact results produced under different thresholds and selecting the setting that best aligns both the real-world dense foot contact and consistency between datasets. 
Specifically, we set the tolerance to 1\,cm for MOYO, 2\,cm for MotionPRO, 3\,cm for PROX and EgoBody, and 5\,cm for BEHAVE and InterCap. 
For datasets that already provide ground-truth dense body contact labels, such as Hi4D and RICH, we directly adopt the provided contact annotations. 
This unified procedure equips all datasets, regardless of whether they include explicit scene meshes, with dense per-vertex foot contact labels that are geometrically consistent. 
The ground plane configurations and ground plane parameters used and extracted in this process are each summarized in Table~\ref{tab:ground_mesh} and Table~\ref{tab:ground_coefficient}.

\section{Visualization of ground-aware learning}
\label{sec:supp_vis_ground_aware}
\begin{figure*}[t]
\begin{center}
\includegraphics[width=0.9\linewidth]{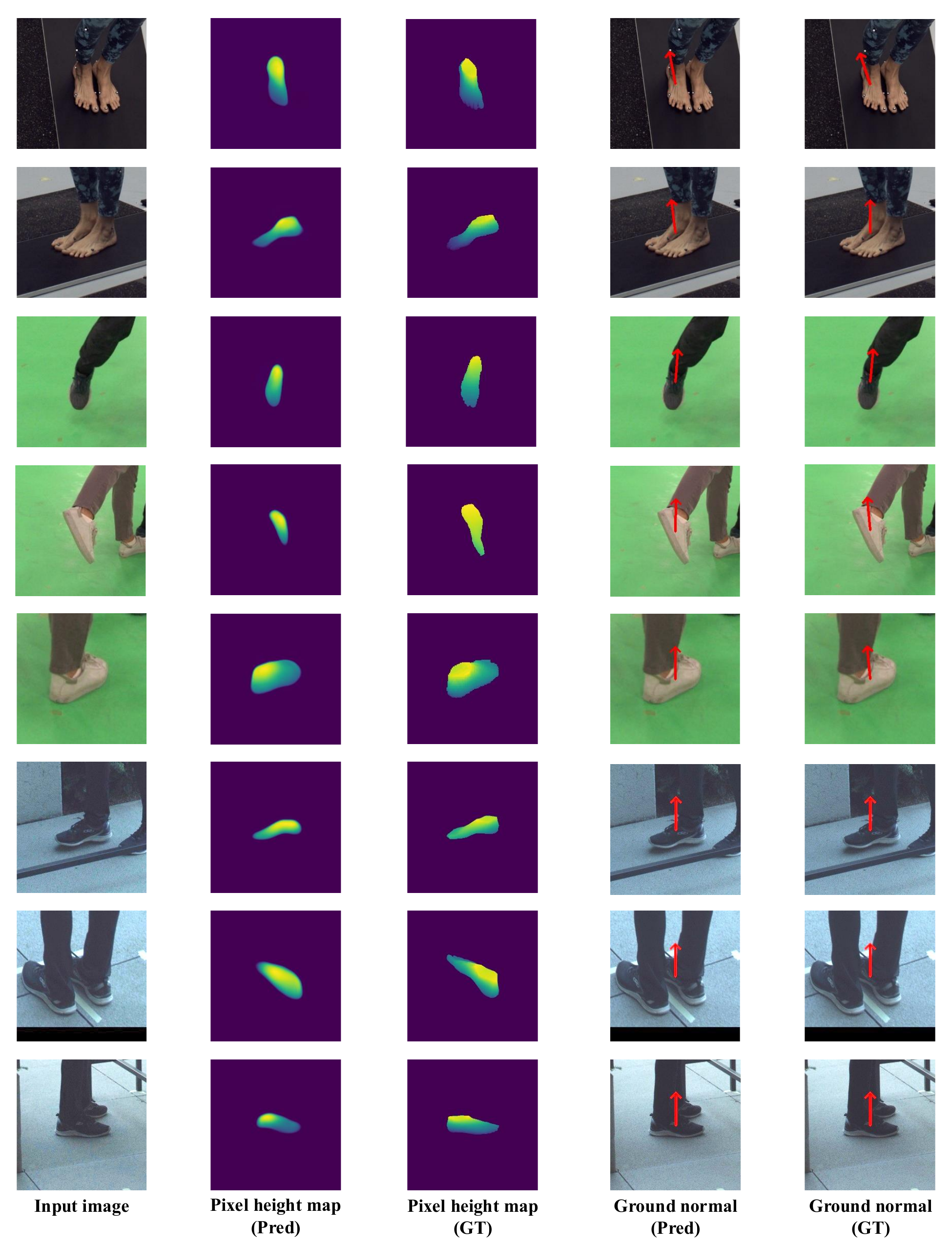}
\end{center}
\vspace{-3.5mm}
\caption{
\textbf{Visualization of ground-aware learning on MOYO~\cite{tripathi20233d}, Hi4D~\cite{yin2023hi4d}, RICH~\cite{huang2022capturing} dataset.}}
\label{fig:supp_ground_aware_qual}
\vspace{-0.3cm}
\end{figure*}

Figure~\ref{fig:supp_ground_aware_qual} presents qualitative results of our ground-aware learning module, which predicts pixel height maps and ground normals from the ground feature. 
The predicted pixel height maps produced by FECO closely match the ground truth. 
In particular, the smooth gradient patterns in the pixel height maps indicate that FECO effectively infers per-pixel height relative to the ground for regions corresponding to the foot. 
Although the predictions exhibit minor smoothing and reduced detail around individual toes, the height maps still provide a strong signal and proof of the ground-aware learning.
Furthermore, the predicted ground normals show strong alignment with the corresponding ground-truth normals. 
The predictions remain robust even when the foot is tilted or not aligned with the ground plane, as illustrated in the third and fourth rows of Figure~\ref{fig:supp_ground_aware_qual}. 
These accurate ground-aware representations of pixel height and ground normal enable FECO to achieve reliable and precise dense foot contact estimation.

\section{Quantitative results on more datasets}
\label{sec:supp_quant_more_datasets}
In addition to MMVP~\cite{zhang2024mmvp} dataset, we further validate FECO across diverse datasets to provide additional benchmarks. 
On BEHAVE~\cite{bhatnagar2022behave}, FECO achieves strong results on all metrics with an F1-score of 0.795, demonstrating robust performance on foot--object contact estimation. 
On RICH~\cite{huang2022capturing}, which contains dense foot--scene interactions, FECO obtains reasonable performance in diverse 3D scene. 
We also evaluate on MOYO~\cite{tripathi20233d}, a motion capture dataset designed for extreme yoga poses. 
Due to the strong out-of-distribution poses and limited scene contact supervision, performance drops compared to other datasets. 
Nevertheless, FECO still detects plausible contacts under such challenging conditions. 
Finally, on Hi4D~\cite{yin2023hi4d}, which includes human--human interactions, our model achieves strong overall results. 
These findings collectively highlight that FECO not only performs well on MMVP, but also yields reasonable performance across diverse real-world interaction scenarios, including object-centric, extreme-pose, and human interaction settings.

\section{Quantitative results on different backbones}
\label{sec:supp_diff_backbones}
We evaluate the performance of FECO using various backbone architectures on the MMVP~\cite{zhang2024mmvp} dataset, keeping all other components fixed.
As summarized in Table~\ref{tab:supp_diff_back}, the ViT-H~\cite{dosovitskiy2020image} backbone achieves the highest F1-score of 0.577, demonstrating the strongest overall performance among all tested models.
ViT-based architectures consistently outperform convolutional alternatives, reflecting the benefit of Transformer~\cite{vaswani2017attention}-based designs in capturing long-range dependencies crucial for dense foot contact estimation.
Among the ViT variants, ViT-B attains the highest recall (0.650), suggesting better coverage of subtle contact regions, while ViT-L achieves a balanced trade-off between precision and recall.
ViT-S maintains competitive performance despite its compact size, underscoring the scalability of Transformer backbones under resource constraints.
Among convolutional backbones, ResNet-152~\cite{he2016deep} delivers the strongest result with an F1-score of 0.533, followed by ResNet-101 and ResNet-50.
For these models, we employ a decoder based on the re-implementation of FCN~\cite{long2015fully} from PyTorch~\cite{paszkepytorch} to predict pixel height maps.
To maintain efficiency, dilation is omitted in shallower variants (ResNet-18 and ResNet-34), though this leads to reduced accuracy due to their limited receptive fields.
Overall, these results confirm that ViT backbones are better suited for dense foot contact estimation than convolutional counterparts, primarily due to their global context modeling and superior spatial reasoning.
The strong performance of ViT-H further justifies its selection as the default backbone in our main experiments.
All model variants will be publicly released.

\begin{table}[t]
\centering
\caption{\textbf{Comparison of various backbone models on MMVP~\cite{zhang2024mmvp} dataset.} All backbones are initialized with ImageNet~\cite{deng2009imagenet} dataset.}
\scalebox{0.87}{\begin{tabular}{lccccc} \toprule
Model & Backbone & $\text{Precision}{\uparrow}$ & $\text{Recall}{\uparrow}$ & $\text{F1-Score}{\uparrow}$ \\
\midrule
FECO & ViT-H~\cite{dosovitskiy2020image} & \underline{0.563} & 0.613 & \textbf{0.577} \\
FECO & ViT-L~\cite{dosovitskiy2020image} & \textbf{0.578} & 0.568 & 0.567 \\
FECO & ViT-B~\cite{dosovitskiy2020image} & 0.526 & \textbf{0.650} & \underline{0.574} \\
FECO & ViT-S~\cite{dosovitskiy2020image} & 0.507 & 0.620 & 0.546 \\
\midrule
FECO & ResNet-152~\cite{he2016deep} & 0.463 & \underline{0.647} & 0.533 \\
FECO & ResNet-101~\cite{he2016deep} & 0.530 & 0.550 & 0.526 \\
FECO & ResNet-50~\cite{he2016deep} & 0.497 & 0.579 & 0.515 \\
FECO & ResNet-34~\cite{he2016deep} & 0.448 & 0.550 & 0.481 \\
FECO & ResNet-18~\cite{he2016deep} & 0.484 & 0.458 & 0.437 \\

\bottomrule
\end{tabular}}
\vspace{-0.3cm}
\label{tab:supp_diff_back}
\end{table}

\section{Computational requirements}
\label{sec:supp_computation}
Table~\ref{tab:supp_comp_require} reports the computational requirements of FECO under different backbone configurations. 
Large-scale Vision Transformer backbones~\cite{dosovitskiy2020image}, such as ViT-H, deliver the strongest representational capacity but also incur the highest computational overhead. 
The ViT-H variant requires more than 34 GB of training memory and nearly 6.5 GB of inference memory. 
This overhead primarily arises from maintaining strong gradients across the model’s 964.64M parameters during training.
ViT-L provides a more moderate alternative, reducing both memory usage and computational cost while retaining high model capacity. 
Smaller variants such as ViT-B and ViT-S provide efficient configurations, with training memory under 13 GB and inference speeds exceeding 23 fps, making them well-suited for practical applications.

ResNet-based backbones~\cite{he2016deep} further expand the design space by offering a spectrum of efficient choices. 
ResNet-152 and ResNet-101 provide strong representational power while maintaining stable inference speeds around 20 fps. 
ResNet-50 achieves a favorable balance, combining manageable memory consumption with fast inference exceeding 24 fps. 
Lightweight variants such as ResNet-34 and ResNet-18 are highly efficient, requiring fewer than 5 GB of training memory and achieving speeds above 40 fps with very low GFLOPs, making them ideal for real-time scenarios and deployment on resource-constrained hardware. 

Overall, FECO supports a wide range of backbones, from high-capacity Transformers to lightweight ResNets, enabling its use across diverse downstream tasks that may have varying computational requirements.

\begin{table}[t]
\centering
\small
\setlength{\tabcolsep}{3pt}
\caption{\textbf{Quantitative results on more datasets}}
\begin{tabular}{llccc}
\toprule
Model & Evaluation Dataset & Precision~$\uparrow$ & Recall~$\uparrow$ & F1-Score~$\uparrow$ \\
\midrule
FECO & BEHAVE~\cite{bhatnagar2022behave}
& 0.755 & 0.917 & 0.795 \\
\midrule
FECO & RICH~\cite{huang2022capturing}
& 0.581 & 0.780 & 0.621 \\
\midrule
FECO & MOYO~\cite{tripathi20233d}
& 0.546 & 0.573 & 0.504 \\
\midrule
FECO & Hi4D~\cite{yin2023hi4d}
& 0.761 & 0.888 & 0.796 \\
\bottomrule
\end{tabular}
\label{tab:supp_other_datasets_eval}
\vspace{-0.3cm}
\end{table}

\section{More qualitative results}
\label{sec:supp_more_qual}
\begin{figure*}[t]
\begin{center}
\includegraphics[width=0.9\linewidth]{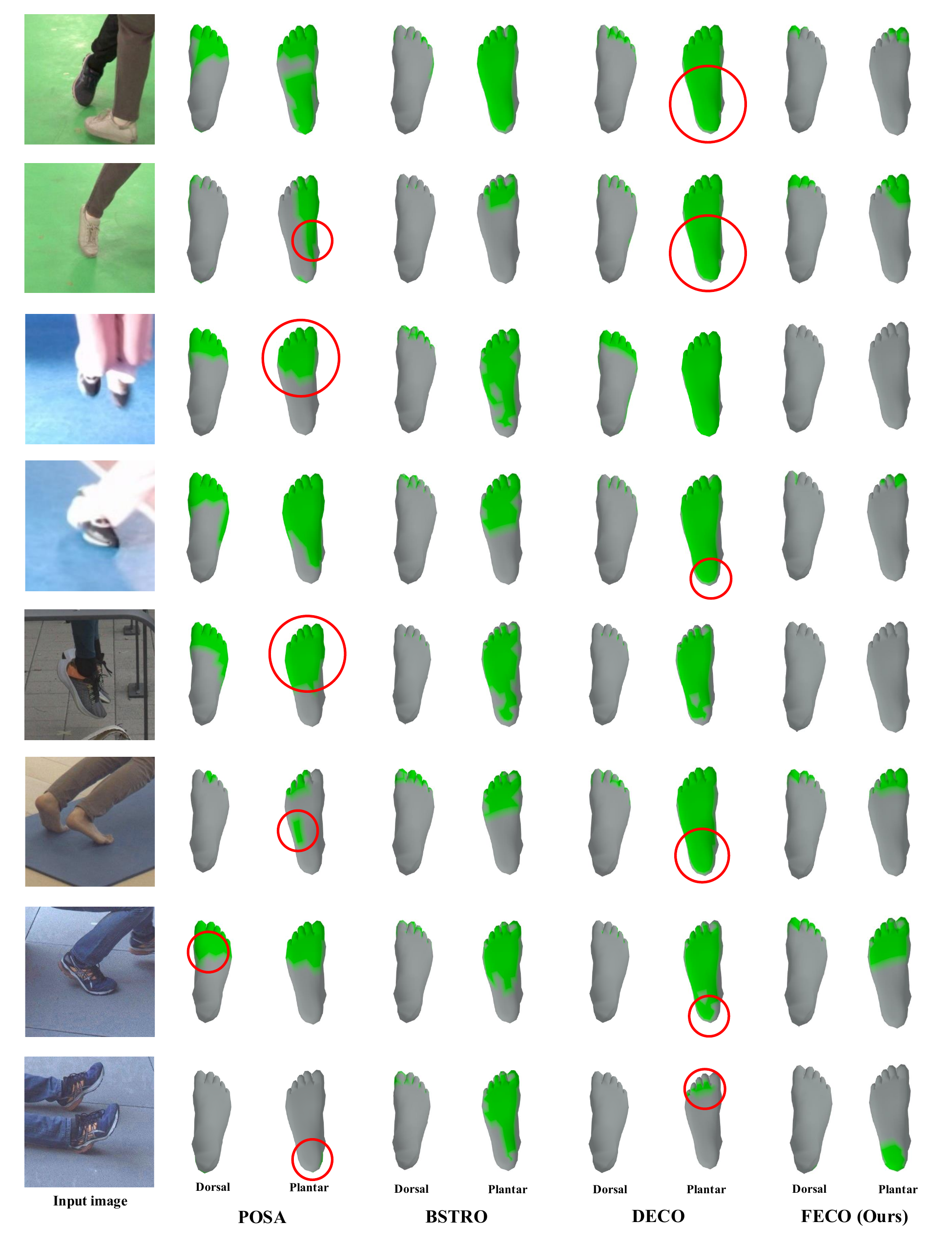}
\end{center}
\vspace{-3.5mm}
\caption{
\textbf{Qualitative comparison of dense foot contact estimation with POSA~\cite{hassan2021populating}, BSTRO~\cite{huang2022capturing}, DECO~\cite{tripathi2023deco} on Hi4D~\cite{yin2023hi4d}, MMVP~\cite{zhang2024mmvp}, RICH~\cite{huang2022capturing} dataset.} Red circles indicate exemplar regions that FECO outperforms previous methods.}
\label{fig:supp_more_qual}
\vspace{-0.3cm}
\end{figure*}

\begin{figure*}[t]
\begin{center}
\includegraphics[width=0.9\linewidth]{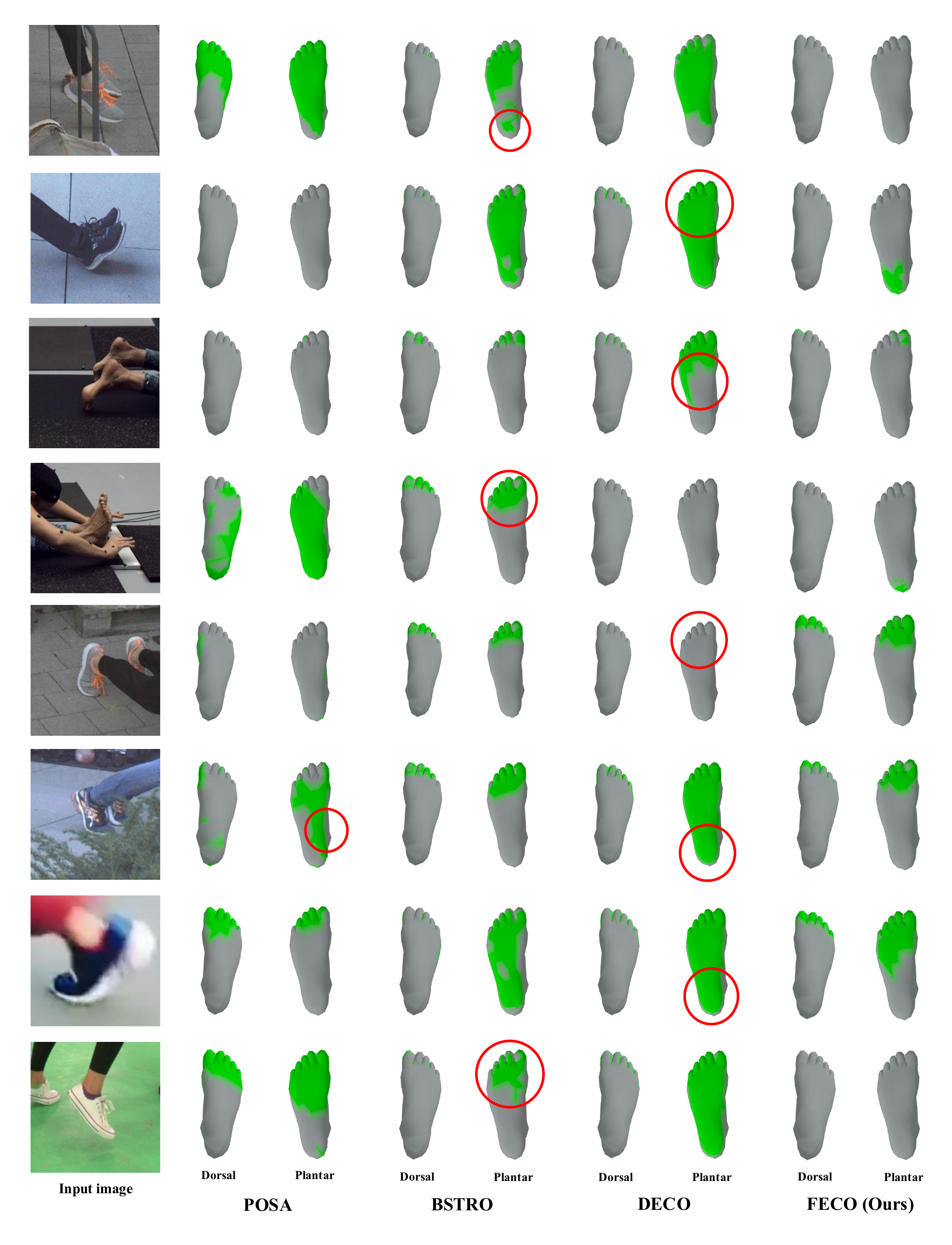}
\end{center}
\vspace{-3.5mm}
\caption{
\textbf{Qualitative comparison of dense foot contact estimation with POSA~\cite{hassan2021populating}, BSTRO~\cite{huang2022capturing}, DECO~\cite{tripathi2023deco} on RICH~\cite{huang2022capturing}, MOYO~\cite{tripathi20233d}, Hi4D~\cite{yin2023hi4d}, COFE dataset.} Red circles indicate exemplar regions that FECO outperforms previous methods.}
\label{fig:supp_more_qual1}
\vspace{-0.3cm}
\end{figure*}

Figure~\ref{fig:supp_more_qual} and Figure~\ref{fig:supp_more_qual1} present additional qualitative comparisons of POSA~\cite{hassan2021populating}, BSTRO~\cite{huang2022capturing}, and DECO~\cite{tripathi2023deco} on the Hi4D~\cite{yin2023hi4d}, MMVP~\cite{zhang2024mmvp}, RICH~\cite{huang2022capturing}, MOYO~\cite{tripathi20233d}, and COFE dataset.
Our FECO consistently outperforms previous state-of-the-art methods by a large margin.
DECO frequently predicts full plantar contact regardless of the actual foot contact observable from the image.
POSA frequently yields false positives contact prediction when the foot pose suggests contact despite there is no contact occurrence in the image.
For example, in the fifth row of Figure~\ref{fig:supp_more_qual}, although the foot does not touch the ground, its pose leads POSA to hallucinate non-existent contact.
BSTRO fails to infer contact in the heel region.
For instance, in the last row of Figure~\ref{fig:supp_more_qual} and the second row of Figure~\ref{fig:supp_more_qual1}, BSTRO does not detect heel contact even though it is well present.
This behavior arises from the absence of ground-aware learning in BSTRO, whereas FECO explicitly incorporates pixel height maps and ground normals to enable ground-aware contact reasoning.
Furthermore, FECO maintains accurate predictions across diverse shoe styles in all samples, which is enabled by its shoe style-invariant learning.

\section{Limitations and societal impacts}
\label{sec:limit_and_impacts}
\noindent\textbf{Limitations.}
Our FECO overcomes the challenge of diverse shoe appearance, which is inherent and unique problem for dense foot contact estimation.
Also, to overcome the issue of lack of in-the-wild training and evaluation dataset for foot contact, we manually annotate and introduce COFE dataset.
However, we mainly operate under foot cropped image, which may not provide useful information when fully occluded.
Advancing towards the integration with dense full-body contact estimation methods or proposing additional module that takes information from full image without being biased with full body pose would solve such occlusion problem.
Also, dense foot contact can be much easier problem with temporal information as static property of foot during contact is, should not serve as a sole contributor but, important cue for contactness.
Developing video-based~\cite{nam2023cyclic} dense foot contact estimation would greatly improve the performance of the model.
Lastly, modeling interactions between the two feet analogous to two hands interaction modeling~\cite{park2023extract} can be a promising direction for future research.

\noindent\textbf{Societal impacts.}
The proposed method has broad potential for applications in sports analytics, rehabilitation, AR/VR, and human behavior understanding.
Nevertheless, deploying dense foot contact estimation in the wild entails risks related to privacy, safety, and sustainability.
Patterns of foot contact during daily activity may reveal aspects of health, so any data collection should occur only with consent, minimal retention, and, when possible, on-device processing.
Our method is not designed as a diagnostic tool, and such use would require higher accuracy and fidelity on specific diagnostic scenarios.
To build a diagnostic system on top of FECO, one would need calibrated confidence estimates, detection of out-of-distribution inputs, expert oversight, and an intended-use license that restricts surveillance and other misuse.

\begin{table*}[t]
\centering
\caption{\textbf{Coefficients of fitted ground plane of datasets.} All refers to all sequences in the dataset.}
\vspace{0.6em}
\scalebox{1.0}{\begin{tabular}{lccccc} \toprule
Dataset & Sequence & slope of $g_1$ & slope of $g_2$ & intercept of $h$ \\
\midrule
PROX~\cite{hassan2019resolving} & Quantitative & $x$: 0.005543 & $y$: 0.021068 & $z$: -0.116057 \\
\midrule
\multirow[c]{7}{*}{BEHAVE~\cite{bhatnagar2022behave}} & Date01 & $x$: 0.027887 & $z$: -0.008862 & $y$: 1.201017 \\
 & Date02 & $x$: -0.003986 & $z$: 0.010984 & $y$: 1.178358 \\
  & Date03 & $x$: 0.033936 & $z$: 0.016988 & $y$: 1.230144 \\
  & Date04 & $x$: 0.007792 & $z$: -0.009504 & $y$: 1.194403 \\
  & Date05 & $x$: 0.022923 & $z$: -0.012357 & $y$: 1.193551 \\
  & Date06 & $x$: 0.019536 & $z$: -0.018724 & $y$: 1.213272 \\
  & Date07 & $x$: 0.009183 & $z$: -0.003937 & $y$: 1.217363 \\
\midrule
InterCap~\cite{huang2022intercap} & All & $x$: 0.027502 & $z$: -0.134394 & $y$: 1.454231 \\
\midrule
\multirow[c]{15}{*}{EgoBody~\cite{zhang2022egobody}}
 & seminar\_g110 & $x$: -0.012439 & $z$: 0.002003 & $y$: -1.661606 \\
 & seminar\_h52 & $x$: 0.001277 & $z$: 0.001124 & $y$: -0.505707 \\
 & kitchen\_gfloor & $x$: 0.001684 & $z$: 0.000132 & $y$: -0.842190 \\
 & cab\_e & $x$: 0.000703 & $z$: 0.006315 & $y$: -0.711828 \\
 & cab\_h\_tables & $x$: -0.000753 & $z$: -0.000542 & $y$: -0.331833 \\
 & seminar\_d78 & $x$: 0.002037 & $z$: 0.000250 & $y$: -0.814285 \\
 & seminar\_d78\_0318 & $x$: -0.001632 & $z$: 0.002296 & $y$: -1.023084 \\
 & seminar\_g110\_0415 & $x$: 0.000735 & $z$: -0.001843 & $y$: -0.789173 \\
 & foodlab\_0312 & $x$: -0.005830 & $z$: -0.000577 & $y$: -0.709450 \\
 & cab\_g\_benches & $x$: -0.000604 & $z$: -0.000745 & $y$: -0.155386 \\
 & seminar\_j716 & $x$: 0.000745 & $z$: -0.001081 & $y$: -0.894595 \\
 & seminar\_h53\_0218 & $x$: -0.000041 & $z$: -0.003415 & $y$: -0.786509 \\
 & cnb\_dlab\_0215 & $x$: 0.000525 & $z$: -0.000740 & $y$: -0.119325 \\
 & seminar\_g110\_0315 & $x$: 0.003623 & $z$: -0.000446 & $y$: -0.733775 \\
 & cnb\_dlab\_0225 & $x$: 0.000715 & $z$: 0.001987 & $y$: -0.155290 \\
\midrule
\multirow[c]{5}{*}{RICH~\cite{huang2022capturing}} 
 & BBQ & $x$: 0.028968 & $z$: -0.137393 & $y$: 1.421628 \\
 & Gym & $x$: -0.055151 & $z$: -0.237227 & $y$: 1.577146 \\
 & LectureHall & $x$: 0.027296 & $z$: -0.299544 & $y$: 1.979380 \\
 & ParkingLot1 & $x$: -0.022767 & $z$: -0.128437 & $y$: 1.372453 \\
 & ParkingLot2 & $x$: -0.018011 & $z$: -0.131626 & $y$: 1.375236 \\
\midrule
MOYO~\cite{tripathi20233d} & All & $x$: 0.000000 & $z$: 0.000000 & $y$: 0.000000 \\
\midrule
Hi4D~\cite{yin2023hi4d} & All & $x$: -0.010825 & $z$: 0.011383 & $y$: 0.274179 \\
\midrule
\multirow[c]{11}{*}{MMVP~\cite{zhang2024mmvp}} 
 & S01 & $x$: 0.000000 & $z$: 0.000000 & $y$: 1.253302 \\
 & S02 & $x$: 0.000000 & $z$: 0.000000 & $y$: 1.245441 \\
 & S03 & $x$: 0.000000 & $z$: 0.000000 & $y$: 1.242629 \\
 & S04 & $x$: 0.000000 & $z$: 0.000000 & $y$: 1.247209 \\
 & S05 & $x$: 0.000000 & $z$: 0.000000 & $y$: 1.219583 \\
 & S06 & $x$: 0.000000 & $z$: 0.000000 & $y$: 1.254879 \\
 & S07 & $x$: 0.000000 & $z$: 0.000000 & $y$: 1.255879 \\
 & S09 & $x$: 0.000000 & $z$: 0.000000 & $y$: 1.240000 \\
 & S10 & $x$: 0.000000 & $z$: 0.000000 & $y$: 1.266970 \\
 & S11 & $x$: 0.000000 & $z$: 0.000000 & $y$: 1.223758 \\
 & S12 & $x$: 0.000000 & $z$: 0.000000 & $y$: 1.246929 \\
\midrule
MotionPRO~\cite{ren2025motionpro} & All & $x$: 0.000000 & $z$: 0.000000 & $y$: 0.000000 \\

\bottomrule
\end{tabular}}
\vspace{-0.3cm}
\label{tab:ground_coefficient}
\end{table*}

\clearpage

{
    \small
    \bibliographystyle{ieeenat_fullname}
    \bibliography{main}
}

\end{document}